\documentclass{article}
\usepackage{amssymb}

\usepackage[final]{corl_2026} 
\usepackage{microtype}
\usepackage{amsmath, amssymb}
\usepackage{graphicx}
\usepackage{stfloats}
\usepackage{booktabs, multirow}
\usepackage{balance}
\usepackage{float}
\usepackage{wrapfig}
\usepackage{caption}
\usepackage{enumitem}
\newcommand{\bl}[1]{{\flushleft\textbf{#1}}}
\newcommand{\meanstd}[2]{$#1${\tiny$\;\pm\,#2$}}

\usepackage[table]{xcolor}
\usepackage{nicematrix}
\usepackage[normalem]{ulem}

\definecolor{Mahogany}{RGB}{192,64,0}
\definecolor{Bittersweet}{RGB}{254,111,94}
\definecolor{RoyalBlue}{RGB}{65,105,225}
\definecolor{BrickRed}{RGB}{203,65,84}
\definecolor{Goldenrod}{RGB}{218,165,32}
\definecolor{OliveGreen}{RGB}{85,107,47}
\definecolor{Fuchsia}{RGB}{255,0,255}
\definecolor{White}{RGB}{255,255,255}

\definecolor{dusty}{RGB}{228,231,235}   
\definecolor{beige}{RGB}{245,238,230}   
\definecolor{bluegray}{RGB}{235,240,247} 

\definecolor{AccentBlue}{RGB}{20,90,160}
\newcommand{\mbe}[1]{\textcolor{AccentBlue}{\textbf{#1}}}

\title{Scaling Sim-to-Real VLA Reinforcement \\ Learning with Generative 3D Worlds \vspace{-0.4cm}}

\author{
    Andrew Choi, Xinjie Wang, Zhizhong Su, and Wei Xu$^{\dagger}$ \\
    Horizon Robotics \ $^\dagger$Corresponding author \\
    \texttt{\footnotesize https://horizonrobotics.github.io/gail/scale-vla-rl}
}

\begin{document}
\maketitle

\vspace{-0.9cm}

\begin{figure}[h]
\centering
\includegraphics[width=\columnwidth]{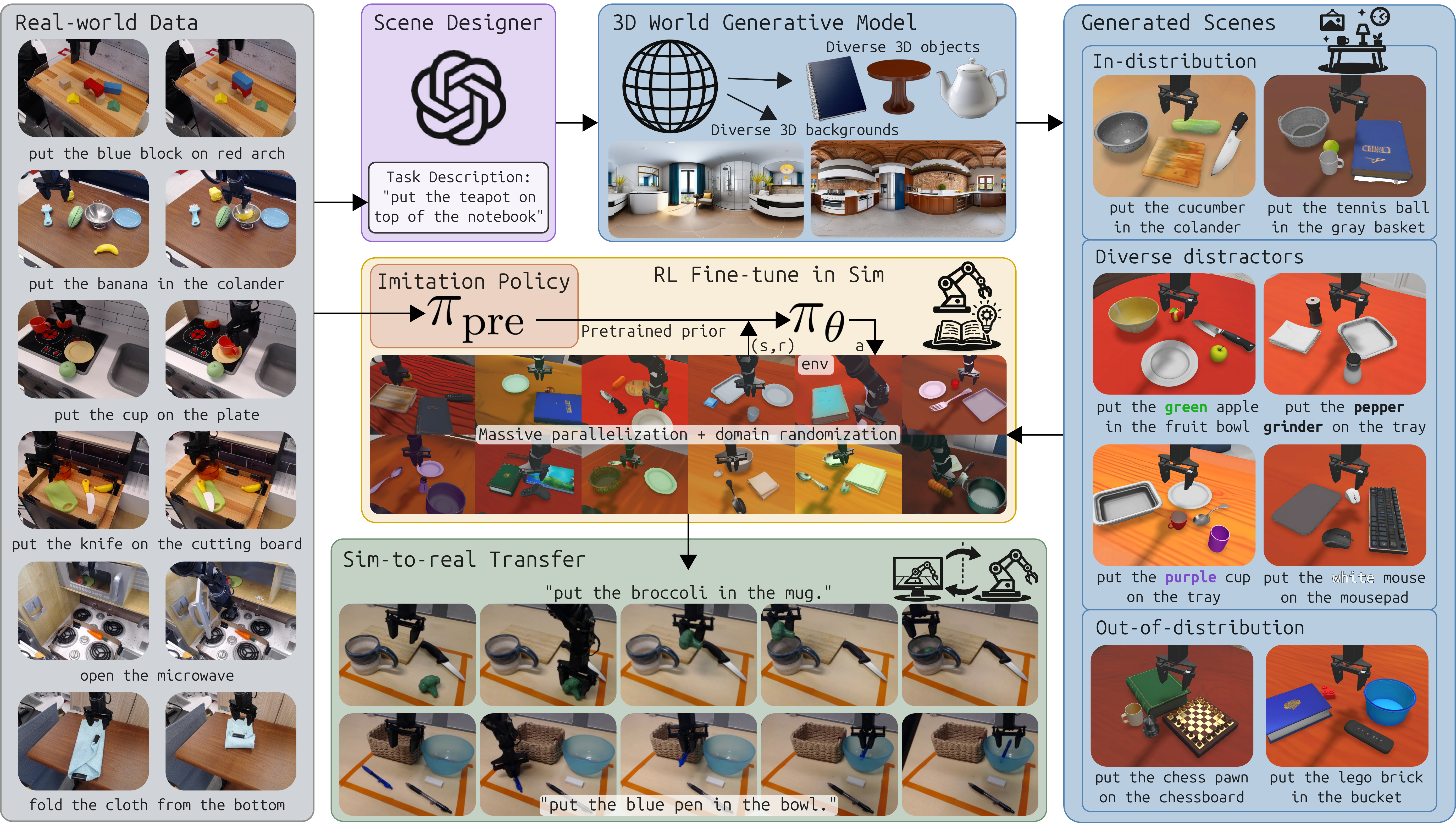}
\caption{
Overall pipeline diagram. Real-world data trains an imitation policy $\pi_\textrm{pre}$. Task descriptions are fed to a language-driven scene designer, which forwards layouts to a 3D world generative model to produce fully generated scenes. $\pi_\theta$ is trained across these scenes, initialized from $\pi_\textrm{pre}$, with massive parallelization and domain randomization. Finally, the trained $\pi_\theta$ is deployed in the real world.
}
\label{fig:front_page}
\end{figure}


\vspace{-0.3cm}
\begin{abstract}
    The strong performance of large vision–language models (VLMs) trained with reinforcement learning (RL) has motivated similar approaches for fine-tuning vision–language–action (VLA) models in robotics.
    Many recent works fine-tune VLAs directly in the real world to avoid addressing the sim-to-real gap.
    While real-world RL circumvents sim-to-real issues, it inherently limits the generality of the resulting VLA, as scaling scene and object diversity in the physical world is prohibitively difficult.
    This leads to the paradoxical outcome of transforming a broadly pretrained model into an overfitted, scene-specific policy.
    Training in simulation can instead provide access to diverse scenes, but designing those scenes is also costly.
    In this work, we show that VLAs can be RL fine-tuned across broad scene and object distributions and with reduced labor by leveraging 3D world generative models.
    Using these models together with a language-driven scene designer, we generate 100 diverse interactive scenes containing unique objects and backgrounds, enabling scalable and highly parallel policy learning.
    Starting from a pretrained imitation baseline, our approach increases simulation success from 9.7\% up to 79.8\% while achieving a 1.25$\times$ speedup in task completion time.
    We further demonstrate successful sim-to-real transfer enabled by the quality of the generated scenes together with domain randomization, improving real-world success from 21.7\% to 75\% and achieving a 1.13$\times$ speedup.
    Finally, we further highlight the benefits of leveraging the effectively unlimited data from 3D world generative models through an ablation study showing that increasing scene diversity directly improves zero-shot generalization.
    Video results are available at our \href{https://horizonrobotics.github.io/gail/scale-vla-rl}{project page}.
\end{abstract}


\vspace{-10pt}
\section{Introduction}
Though internet-scale training of VLMs has seen explosive success, the performance of VLAs has comparatively lagged behind.
As of recently, the dominant pipeline for training robot foundation models consists of leveraging all layers of the ``data pyramid'': 1) pretraining a VLM backbone on internet-scale data, 2) further pretraining via imitation learning on large-scale robot datasets~\citep{open_x_embodiment_rt_x_2023, deng2025graspvlagraspingfoundationmodel}, and 3) task- and embodiment-specific fine-tuning via imitation learning, real-world reinforcement learning, or sim-to-real reinforcement learning.
While the first two stages are largely addressed by training on publicly available data, the third stage often requires significant human involvement.

For imitation learning and offline RL, this involvement comes in the form of manual real-world data collection.
For real-world RL, human-in-the-loop (HIL) supervision is often required to achieve sufficient sample efficiency~\citep{chen2025conrft, intelligence2025pi06vlalearnsexperience}.
Conversely, sim-to-real RL eliminates the need for manual data collection and HIL by instead generating an abundance of data through massive parallelization~\citep{li2026simplevlarl}, but introduces new challenges in the form of visual and dynamic sim-to-real gaps.
Furthermore, though elements of the learning pipeline such as environment resets and reward detection are significantly easier in simulation, designing the 3D environments themselves---albeit a one-time cost---also requires substantial human effort.

An often-overlooked drawback of real-world fine-tuning is the loss of generality that arises when training on only a small number of scenes.
Because scaling scene diversity in the physical world is prohibitively expensive, most prior work fine-tunes VLAs in narrowly scoped settings.
As a result, fine-tuning frequently transforms a broadly pretrained VLA into an overfitted, scene-specific policy.
In this paper, we show that generative 3D world models can automatically construct large training distributions of interactive environments for RL fine-tuning of VLAs, enabling scalable sim-to-real learning beyond the small number of hand-designed environments used in prior work.
A snapshot of the overall pipeline can be seen in Fig.~\ref{fig:front_page}.
Overall, our main contributions involve showing that
\begin{enumerate}[leftmargin=12pt, topsep=2pt, itemsep=2pt]
    \item \textbf{Reinforcement learning and 3D world generative models strongly complement the large capacity of VLAs.}
    We introduce a language-driven scene designer that converts task descriptions into structured scene layouts, which are then used by a 3D world generative model to produce fully interactive environments.
    Paired with highly parallelized RL, this enables VLAs to be fine-tuned across a wide variety of scenes without requiring manual scene design.
    In this work, we fine-tune a pretrained VLA in simulation, achieving up to a \mbe{70.1-percentage-point improvement in success rate and a 1.25$\times$ task completion speedup} across 100 unique generated scenes.
    \item \textbf{Increasing scene diversity substantially improves zero-shot generalization.} 
    Through extensive ablations in simulation, we show that zero-shot generalization scales positively with training distribution diversity. 
    Results show that fine-tuning on 50 scenes can \mbe{increase zero-shot success rate by 24.7 percentage points} compared to fine-tuning on a single scene.
    \item \textbf{Techniques for sim-to-real transfer of shallow models carry over to VLAs.}
    We achieve strong sim-to-real deployment results by leveraging 1) high-fidelity generated 3D objects and scenes, 2) simple domain randomization, and 3) using PD control with gravity compensation.
    We achieve a \mbe{53.3-percentage-point improvement in success rate and a 1.13$\times$ task completion speedup} across 12 scenes and 240 real-world experiments.
    \item \textbf{RL fine-tuning flow matching VLAs into Gaussian policies is effective.} 
    We demonstrate that PPOFlow—a PPO-based~\citep{schulman2017proximalpolicyoptimizationalgorithms} variant of ReinFlow~\citep{zhang2025reinflow}—effectively fine-tunes large pretrained flow matching VLAs.
    PPOFlow can transform a multi-step flow matching policy into a single-step Gaussian policy, which was previously thought to harm performance~\citep{dppo2024}.
    Instead, our findings directly support the recent work of~\citet{pan2026much} in that the multimodality of diffusion models is not an important factor for having a high-performing robot policy.
    Furthermore, we demonstrate an \mbe{inference latency speedup of 2.36$\times$} through PPOFlow by removing the need for iterative denoising of actions, similar to consistency policies~\citep{chen2024boostingcontinuouscontrolconsistency, prasad2024consistencypolicyacceleratedvisuomotor, chen2025conrft}.
    More details can be seen in Sec.~\ref{sec:k_ablation}.
\end{enumerate}

\section{Related Work}

\bl{VLA models.} 
Motivated by the successes of VLMs in computer vision and NLP, large open-source robot datasets, both physical~\citep{walke2023bridgedata, open_x_embodiment_rt_x_2023} and synthetic~\citep{deng2025graspvlagraspingfoundationmodel, chen2025robotwin}, have been created to facilitate supervised pretraining of VLAs.
Using such datasets, a series of open-source robot VLAs have been released, such as OpenVLA~\citep{kim2025openvla}, SpatialVLA~\citep{qu2025spatialvlaexploringspatialrepresentations}, GraspVLA~\citep{deng2025graspvlagraspingfoundationmodel}, $\pi_0$~\citep{black2024pi0visionlanguageactionflowmodel}, $\pi_{0.5}$~\citep{black2025pi0.5}, and many more.
A common strategy is to then supervised fine-tune a pretrained model with imitation learning on task-specific demonstration data~\citep{kim2025finetuningvisionlanguageactionmodelsoptimizing}.
Though straightforward, such approaches often suffer in performance when encountering out-of-distribution states~\citep{liu2026rlbringvlageneralization} and recent work has suggested that dataset fragmentation introduces unwanted shortcut learning of VLAs~\citep{xing2025shortcutlearninggeneralistrobot}.
Given this, there has been an increasing trend of instead using reinforcement learning to fine-tune VLAs.

\bl{Learning in real.}
One popular strategy for RL fine-tuning VLAs has been to perform RL directly in the real world~\citep{chen2025conrft,sharma2023selfimproving, intelligence2025pi06vlalearnsexperience, ghasemipour2025selfimproving, yang2024robofume, pmlr-v205-walke23a, mendonca2024continuously}.
Though RL in real circumvents issues arising from sim-to-real gap, such methods must handle tedious issues that don't arise in simulation.
One key consideration is how to administer the reward signal.
For simplicity, most methods often rely on a sparse success reward signal, which is then administered by either a trained reward model~\citep{sharma2023selfimproving, yang2024robofume, chen2025conrft, du2023visionlanguagemodelssuccessdetectors}, human engineered checkers~\citep{pmlr-v205-walke23a}, or a human labeler~\citep{intelligence2025pi06vlalearnsexperience}.
Some approaches have used more dense reward formulations such as steps-to-go~\citep{ghasemipour2025selfimproving}.
Another consideration is how to perform scene resets, which can be handled by manual resets~\citep{chen2025conrft, intelligence2025pi06vlalearnsexperience}, scripted resets~\citep{chen2025conrft, mendonca2024continuously}, or learning the reset task~\citep{sharma2023selfimproving, yang2024robofume, pmlr-v205-walke23a}.
Lastly, it has become increasingly evident that a combination of offline RL and human-in-the-loop learning is a key necessity to get reasonable sample efficiency when executing RL on real hardware~\citep{yang2024robofume, chen2025conrft, intelligence2025pi06vlalearnsexperience}.
Though such methods have shown impressive results for learning tasks on a set scene and objects in a reasonable amount of time, the ability to maintain the generality of such policies in both scene and object space through scaling is a major question, especially given the necessity for human guidance.

\bl{Learning in simulation.}
The other method of RL fine-tuning VLAs is to do so in simulation~\citep{fang2025rebotscalingrobotlearning, li2025vlarftvisionlanguageactionreinforcementfinetuning, li2026simplevlarl}.
Compared to real, training in simulation offers numerous benefits, such as oracle success detection, automatic resets, and access to privileged information for teacher-student training setups~\citep{zhang2025slim, singh2025dextrahrgbvisuomotorpoliciesgrasp, dan2025xsim, chen2025clutterdexgrasp}.
Due to the domain difference, special consideration must be taken to ensure that such policies transfer from sim-to-real successfully.
Several strategies for handling the sim-to-real gap exist including domain randomization~\citep{zhang2025slim, he2025viralvisualsimtorealscale}, real-to-sim delta action models~\citep{he2025asapaligningsimulationrealworld}, real-to-sim physical property estimation~\citep{wang2025phys2realfusingvlmpriors}, real-to-sim scene reconstruction~\citep{sun2025prismprojectionbasedrewardintegration, dan2025xsim, lum2025crossing}, and more (albeit many of these methods have been used on smaller models, rather than foundation-scale).
Training shallow models from scratch often requires heavily engineered reward functions~\citep{zhang2025slim, he2025asapaligningsimulationrealworld, he2025viralvisualsimtorealscale}.
With this, fine-tuning VLAs has immense benefits over shallow models as a well-trained VLA prior only requires a sparse reward and massive parallelization to learn~\citep{li2026simplevlarl}.
Recent works for RL fine-tuning VLAs include leveraging real-to-sim-to-real robot video synthesis for leveraging both the real data and scalability of simulation~\citep{fang2025rebotscalingrobotlearning} as well as learning within learned world models~\citep{li2025vlarftvisionlanguageactionreinforcementfinetuning}.
In contrast to these prior works, our main focus in this manuscript is to instead explore a different angle: can RL fine-tuning be done for as wide of a scene distribution as possible? And if so, how does increasing scene diversity improve zero-shot generalization?

\bl{3D world generative models.} Constructing large-scale, realistic 3D simulation environments remains a fundamental bottleneck for sim-to-real reinforcement learning. Existing 3D generative paradigms (e.g., TRELLIS~\cite{xiang2025structured3dlatentsscalable}, WorldGen~\cite{worldgen2025ziyangxie}) primarily focus on producing static visual assets that lack physical interactivity. Although recent works such as Holodeck~\cite{yang2024holodecklanguageguidedgeneration} and RoboGen~\cite{wang2024robogenunleashinginfinitedata} attempt to procedurally construct interactive environments, end-to-end generation of task-level physical scenes remains challenging. Recently, emerging generative engines, exemplified by EmbodiedGen~\cite{wang2025embodiedgengenerative3dworld, wang2026embodiedgenv2agenticsimulationready}, have enabled natural language-driven generation of interactive 3D scenes. 
In this work, we explore the integration of generative 3D engines as a low-cost, highly parallelizable pipeline for scalable scene generation, enabling large-scale RL fine-tuning of VLA models in simulation.

\section{Methodology}

\begin{figure}[t]
\centering
\includegraphics[width=\columnwidth]{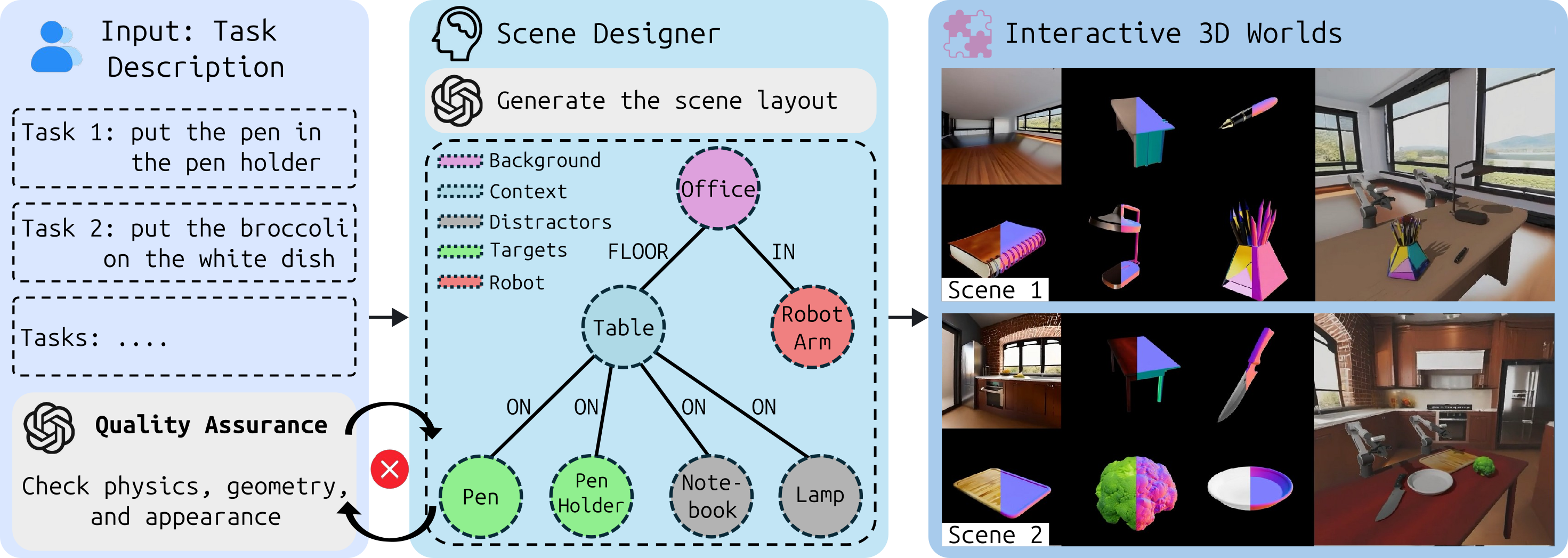}
\caption{Overview of the simulation environment generation pipeline. 
A GPT-4o-powered scene designer converts task descriptions into 
structured scene graphs over semantic roles and spatial relations, 
which are instantiated into fully interactive 3D worlds. 
A quality assurance loop filters physically implausible configurations 
before simulation, enabling scalable, on-demand generation of diverse 
environments for RL training.
}
\label{fig:embodiedgen_layout}
\end{figure}

\bl{Generative simulation.} To construct large-scale, physically interactive 3D environments for sim-to-real reinforcement learning, we employ ManiSkill 3~\citep{taomaniskill3} as our GPU-parallelized simulator and extend EmbodiedGen~\citep{wang2025embodiedgengenerative3dworld, wang2026embodiedgenv2agenticsimulationready} into a comprehensive simulation environment generation backend.
As illustrated in Fig.~\ref{fig:embodiedgen_layout}, EmbodiedGen takes 
a structured scene tree layout as input and produces a fully interactive physical 3D world via its text-to-3D asset generation and layout composition interfaces. 
We develop a scene designer powered by GPT-4o~\citep{achiam2023gpt} that automatically produces valid scene layouts from task descriptions. 
Given an instruction such as ``put the pen in the pen holder'', the scene designer parses it into a structured scene graph comprising core semantic roles (background, context, distractors, manipulation targets, and the robot) along with their spatial relations (e.g., ON, IN). 
This graph-based representation enables flexible, compositional control over scene complexity and distractor density, allowing systematic modulation of training difficulty. 
Each generated scene is then passed through an automated GPT-4o-driven quality assurance loop that checks physical plausibility and geometric consistency, discarding configurations that would cause simulation instability.

\bl{Markov decision process.}
We formulate our learning problem as a finite-horizon partially observable Markov decision process (POMDP) with state space $\mathcal S$, action space $\mathcal A$, and discount factor $\gamma \in [0,1]$.
At each timestep $t$, the agent receives an observation $\mathbf o_t$ and selects an action $\mathbf a_t$ according to a policy $\pi(\mathbf a_t \mid \mathbf o_t)$.
An observation consists of an RGB image $\mathbf I \in \mathbb R^{H \times W \times 3}$, language instruction $\mathbf e \in \mathbb R^{L}$, and proprioceptive information in the form of (xyz, rpy, gripper) end-effector pose $\mathbf q \in \mathbb R^{7}$. 
This full observation is given by $\mathbf o_t = [\mathbf I_t, \mathbf e_t, \mathbf q_t]$.
Actions are represented as (end-effector delta-pose, binary gripper) action chunks $\mathbf A \in \mathbb R^{C \times 7}$, where $C$ denotes the action chunk length.
We treat each action chunk as a single decision step, resulting in a continuous action space $\mathcal A \subset \mathbb R^{C \times 7}$.
The objective is to learn a policy that maximizes the expected discounted return $J(\pi) = \mathbb E_{\pi}[\sum_{t=0}^{T-1} \gamma^t r_t ]$, where $T$ denotes the episode horizon.
We use a sparse rule-based success reward computed directly from simulator states and object relations (Eq.~\ref{eq:reward}).

\begin{figure}[t]
\centering
\includegraphics[width=0.9\columnwidth]{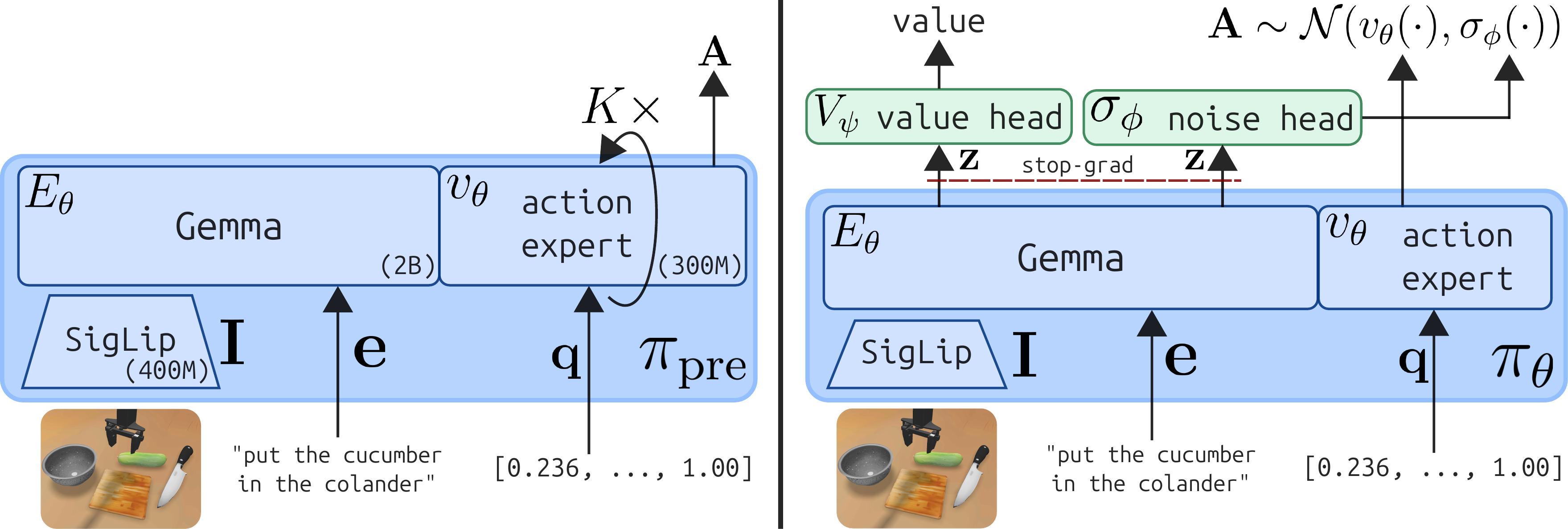}
\caption{
Architecture diagram of the $\pi_0$ model as a pretrained imitation model $\pi_\textrm{pre}$ (left) and then modified for RL fine-tuning, $\pi_\theta$ (right).
}
\label{fig:architecture}
\end{figure}

\bl{Flow matching models.}
For RL fine-tuning to work with solely a sparse reward signal, we look to use a pretrained robot foundation model to boost sample efficiency~\citep{li2026simplevlarl}. 
In this work, we use $\pi_0$~\citep{black2024pi0visionlanguageactionflowmodel} as our baseline imitation model and pretrain it on BridgeV2~\citep{walke2023bridgedata}.
The $\pi_0$ model consists of a VLM backbone $E_{\theta}$ and an action expert head $v_\theta$ (left of Fig.~\ref{fig:architecture}).
It is trained using a rectified flow-matching objective
\begin{align}
    \mathcal L_\textrm{flow}(\theta)
    &=
    \mathbb E_{\mathbf o_t,\mathbf A_t^1 \sim \mathcal D,\boldsymbol \epsilon \sim \mathcal N(\mathbf 0, \mathbf I),\ \tau \sim \mathcal U(0, 1)}
    \left[
    \lVert v_\theta(\mathbf A^\tau_t, KV_{\theta}(\mathbf o_t), \tau) - (\mathbf A_t^1 - \boldsymbol \epsilon) \rVert_2^2
    \right],
\end{align}
where $\mathcal D$ denotes the demonstration dataset, $\tau \in [0,1]$ is the continuous integration time, $KV_{\theta}(\mathbf o_t)$ denotes the key-value tensors of $E_{\theta}(\mathbf o_t)$ for cross-attention, and $\mathbf A^\tau_t = \tau \mathbf A^1_t + (1 - \tau) \boldsymbol \epsilon$.
Robot actions are generated by numerically integrating the learned ODE
\begin{align}
    \frac{d \mathbf A^\tau_t}{d\tau} = v_\theta(\mathbf A^\tau_t, KV_{\theta}(\mathbf o_t), \tau),
\end{align}
where $\mathbf A^0_t\sim \mathcal N(\mathbf 0, \mathbf I)$.
We denote the resulting pretrained imitation policy as $\pi_\textrm{pre}$.

\bl{PPOFlow.}
As flow matching models are deterministic, the importance ratio cannot be computed, preventing standard PPO updates.
ReinFlow solves this by injecting learnable noise $\sigma_\phi(\mathbf A^\tau_t, \mathbf z_t, \tau)$ into the numerical integration process, where $\mathbf z_t$ denotes the stop-gradient hidden state $\mathbf z_t = \textrm{sg}(E_{\theta}(\mathbf o_t))$.
This effectively converts each integration step $\mathbf A^{\tau+\Delta \tau}_t = \mathbf A^\tau_t + v_\theta(\mathbf A^\tau_t, KV_\theta(\mathbf o_t), \tau)\Delta\tau$ into a Gaussian sample 
\begin{align}
    &\hat{\mathbf A} = \mathbf A^\tau_t + v_\theta(\mathbf A^\tau_t, KV_{\theta}(\mathbf o_t), \tau)\Delta\tau, \\
    &\mathbf A^{\tau+\Delta\tau}_t \sim \mathcal N( \hat{\mathbf A}, \sigma_\phi(\mathbf A^\tau_t, \mathbf z_t, \tau)).
\end{align}
Let us denote the number of numerical integration steps as $K = 1 / \Delta \tau$.
This then allows us to compute the joint log probability of the denoising process for a particular step $t$ as
\begin{align}
    \log \pi (\mathbf A_t^0, ..., \mathbf A_t^1 | \mathbf o_t) = \log \mathcal N(\mathbf 0, \mathbf I) + \sum_{k=0}^{K-1} \log \pi \left(\mathbf A_t^{(k+1)\Delta \tau} | \mathbf A_t^{k\Delta \tau}, \mathbf o_t \right).
\label{eq:log_prob}
\end{align}
In addition to the noise head $\sigma_\phi$, we also add a value head $V_\psi(\mathbf z_t)$ for estimating the state value.
A full diagram of $\pi_\theta$ can be observed on the right side of Fig.~\ref{fig:architecture}.
We can then directly use the log probability (Eq.~\ref{eq:log_prob}) to compute the following power-scaled importance ratio, which is then used along with the value from $V_\psi$ in the original PPO clipped objective
\begin{align}
    \hat r_t &= \left(\dfrac{\pi_\theta (\mathbf A^0_t, ..., \mathbf A^1_t | \mathbf o_t)}{\pi_{\theta,\textrm{old}} (\mathbf A^0_t, ..., \mathbf A^1_t | \mathbf o_t)}\right)^s, \label{eq:importance_ratio} \\
    \mathcal L_\textrm{PPOFlow}(\theta, \phi) = \mathbb E_t& \left[ \min \left( \hat r_t \hat A_t, \textrm{clip}(\hat r_t, 1- \epsilon, 1+\epsilon) \hat A_t \right) \right],
\end{align}
where $\hat A_t$ is the advantage estimate computed with GAE~\citep{schulman2018highdimensionalcontinuouscontrolusing}, $\epsilon$ is the clipping range, and $s \in (0, 1]$ is a scaling factor that reduces variance and helps enforce the importance ratio to stay within a more stable numerical range~\citep{zheng2025groupsequencepolicyoptimization}.

\section{Experiments}

\bl{Experiment design.} To study the effect of scaling the number of scenes $N$, we generate 100 unique scenes and denote this full set as $\mathcal{W}$ (see Sec.~\ref{sec:scene_designer} for details and metrics).
We define the axes of scene diversity as variation in background environments and object assets. 
Variations in camera pose, robot pose, object pose, and lighting are treated separately as domain randomization rather than scene diversity. 
Consequently, scaling $N$ primarily measures the effect of broadening the distribution of backgrounds and objects encountered during RL fine-tuning.
In this work, we focus on a pick-and-place task in which, given a language command, the policy must grasp an object and place it on top of another amid several distractors.
\begin{wrapfigure}{r}{0.30\textwidth}
\includegraphics[width=0.30\columnwidth]{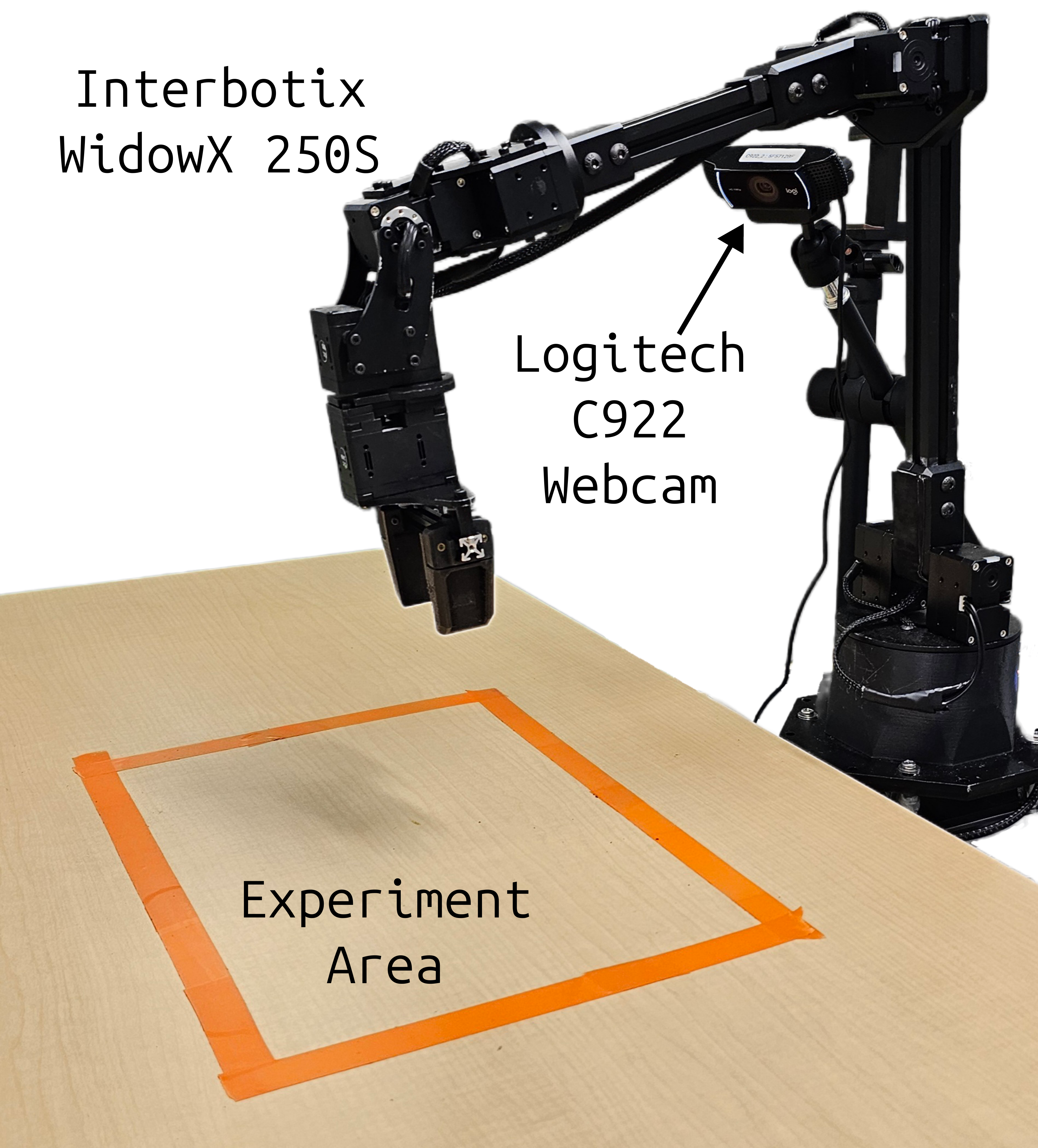}
\caption{Experiment setup.}
\label{fig:robot_fig}
\end{wrapfigure}
To evaluate out-of-distribution (OOD) generalization, we construct three randomly sampled subsets of $\mathcal{W}$, each containing 50 scenes, denoted $\mathcal{H}_i$ for $i \in \{0,1,2\}$.
For each subset, the corresponding OOD set is defined as $\bar{\mathcal{H}}_i = \mathcal{W} \setminus \mathcal{H}_i$.
Using each $\mathcal{H}_i$, we train three independent runs for each $N \in \{1,3,10,25,50\}$, where training uses the first $N$ scenes under a fixed ordering of $\mathcal{H}_i$.
In addition, we train $N=100$ runs on the full set $\mathcal{W}$.
Since $N=100$ necessarily uses the full scene set, we instead report three random-seed runs for this setting.
For embodiment, we use an Interbotix WidowX 250S manipulator with a single external Logitech C922 webcam for vision (Fig.~\ref{fig:robot_fig}).
Real-world evaluation is conducted on an NVIDIA RTX 4090 GPU.

\bl{Baselines and external evaluation benchmarks.} 
The initial flow-matching imitation policy $\pi_\textrm{pre}$ serves as a baseline and uses $K=10$ and action chunk size $C=4$.
$\pi_\textrm{pre}$ is trained on BridgeV2, which is predominantly pick-and-place (over 70\%), with the remainder consisting of tasks such as sweeping, towel folding, and drawer opening.
This yields a strong initialization that is well aligned with our task as well as a reference point for measuring the effect of RL fine-tuning.
To investigate the benefits of training on generated scenes, we introduce another baseline in which we RL fine-tune a policy on three manually designed Bridge tabletop scenes from SimplerEnv~\citep{li2024evaluatingrealworldrobotmanipulation}.
We evaluate OOD generalization on these SimplerEnv scenes and two additional manually constructed external benchmarks: (1) 18 scenes using 25 Google Scanned Objects assets~\citep{downs2022googlescannedobjectshighquality}, with repetitions across scenes and 18 assets semantically OOD to $\mathcal W$, and (2) 25 scenes using 100 Hunyuan3D-2.1-generated assets~\citep{hunyuan3d2025hunyuan3d}, with 48 assets semantically OOD.
Neither external asset source is used during RL training, allowing us to evaluate generalization under both asset-source and generator shifts.
Additional details and qualitative examples for each scene set are provided in Secs.~\ref{sec:embodied_gen_scenes}, \ref{sec:simpler_env_scenes}, \ref{sec:scanned_scenes}, and~\ref{sec:third_party_scenes}.

\begin{wrapfigure}{r}{0.45\textwidth}
\vspace{-8pt}
\includegraphics[width=0.45\columnwidth]{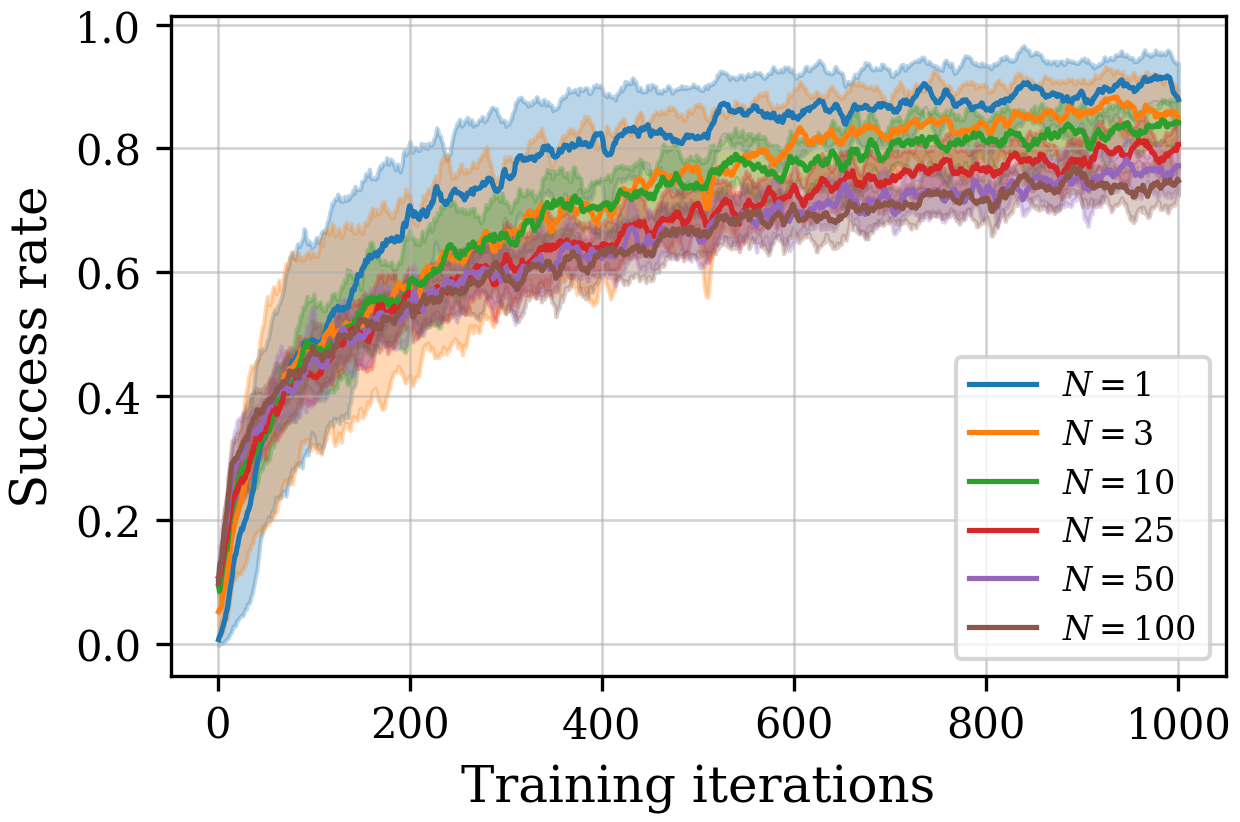}
\caption{Training curves for all $N$.}
\label{fig:train_curves}
\end{wrapfigure}

\bl{Training details.} For training, we use 8 NVIDIA RTX 6000 Ada GPUs to LoRA fine-tune the VLM $E_\theta$ and fully fine-tune the action expert $v_\theta$ over 5 days.
The value head $V_\psi$ and noise head $\sigma_\phi$ are shallow MLPs and are also fully fine-tuned.
Training curves are shown in Fig.~\ref{fig:train_curves}.
To enable successful sim-to-real transfer, we rely on three factors: (1) the high fidelity of generated objects and scenes, 
(2) domain randomization, and (3) PD control with gravity compensation.
With gravity compensation enabled, we found that explicit system identification was largely unnecessary, provided that both simulation and real-world controllers could reach the majority of target poses with sufficiently low tracking error, thereby minimizing the dynamics sim-to-real gap.
We use an action chunk size of $C=4$ and set the number of integration steps to $K=1$.
All other training hyperparameters and domain randomization ranges are provided in Tables~\ref{tab:training_parameters} and~\ref{tab:domain_randomization}, respectively.

\bl{Fixed compute budget.} Note that in Fig.~\ref{fig:train_curves}, the final training success rates decrease monotonically as $N$ increases.
This behavior is expected, as we use the same batch size and mini-batch size for all runs; consequently, each scene receives fewer samples, scaling linearly with $1/N$.
Though computationally impractical, we hypothesize that scaling the batch size and mini-batch size proportionally with $N$ could mitigate this effect.
Importantly, a fixed compute budget disadvantages larger $N$, since the policy must learn over a broader scene distribution while receiving fewer samples from each individual scene.
Despite this reduced per-scene sampling, OOD performance improves substantially with $N$ before saturating around $N=25$--$50$, as shown in the next section.
We hypothesize that beyond $N=25$--$50$, the benefits of additional scene diversity are increasingly offset by reduced per-scene coverage under the fixed compute budget.

\newcolumntype{L}{>{\raggedright\arraybackslash}m{1.55cm}}
\newcolumntype{C}{>{\centering\arraybackslash}m{1.15cm}}
\newcolumntype{G}{>{\centering\arraybackslash}m{0.5cm}}

\begin{table*}[!t]
\renewcommand{\arraystretch}{1.1}
\caption{
Simulation evaluation results.
For a certain policy (row), results are color coded as
{\setlength{\fboxsep}{1pt}\colorbox{beige}{OOD}},
{\setlength{\fboxsep}{1pt}\colorbox{bluegray}{ID}}, and
{\setlength{\fboxsep}{1pt}\colorbox{dusty}{mixture}}.
$\Delta$G = SR$_\mathrm{ID}$$-$SR$_\mathrm{OOD}$
denotes the ID--OOD generalization gap in percentage points;
dashes indicate ID or mixture evaluations, or cases without a defined ID reference.
$^*$ denotes random-seed runs rather than scene-split runs.
}
\centering
\scriptsize

\begin{NiceTabular}{L!{\hspace{-6pt}} CCG CCG CCG}

\toprule
\multirow{2}{*}[-3pt]{Policy}
& \multicolumn{3}{c}{EG ID scenes $\mathcal H_i$ ($N$)}
& \multicolumn{3}{c}{EG OOD scenes $\bar{\mathcal H}_i$ (50)}
& \multicolumn{3}{c}{EG all scenes $\mathcal W$ (100)} \\
\cmidrule(lr){2-4}
\cmidrule(lr){5-7}
\cmidrule(lr){8-10}
& SR (\%) [$\uparrow$] & TF (s) [$\downarrow$] & $\Delta$G
& SR (\%) [$\uparrow$] & TF (s) [$\downarrow$] & $\Delta$G
& SR (\%) [$\uparrow$] & TF (s) [$\downarrow$] & $\Delta$G \\
\midrule

$N=0 \enspace (\pi_\textrm{pre})$
& $-$ & $-$ & $-$
& $9.6$ & $10.3$ & $-$
& $9.7$ & $10.0$ & $-$ \\

$N=3 \enspace (\textrm{SE})$
& $-$ & $-$ & $-$
& \cellcolor{beige} $36.5$
& \cellcolor{beige} $8.5$
& \cellcolor{beige} $60.2$
& \cellcolor{beige} $36.0$
& \cellcolor{beige} $8.6$
& \cellcolor{beige} $60.7$ \\

$N=1$
& \cellcolor{bluegray} \meanstd{\mathbf{94.3}}{2.6}
& \cellcolor{bluegray} \meanstd{\uline{7.7}}{0.9}
& $-$
& \cellcolor{beige} \meanstd{53.2}{5.5}
& \cellcolor{beige} \meanstd{9.0}{0.5}
& \cellcolor{beige} $41.1$
& \cellcolor{dusty} \meanstd{51.6}{5.3}
& \cellcolor{dusty} \meanstd{9.0}{0.3}
& $-$ \\

$N=3$
& \cellcolor{bluegray} \meanstd{\uline{88.7}}{4.7}
& \cellcolor{bluegray} \meanstd{\mathbf{7.6}}{0.5}
& $-$
& \cellcolor{beige} \meanstd{61.3}{1.8}
& \cellcolor{beige} \meanstd{8.7}{0.5}
& \cellcolor{beige} $27.4$
& \cellcolor{dusty} \meanstd{60.0}{4.6}
& \cellcolor{dusty} \meanstd{8.8}{0.5}
& $-$ \\

$N=10$
& \cellcolor{bluegray} \meanstd{87.0}{3.3}
& \cellcolor{bluegray} \meanstd{\uline{7.7}}{0.4}
& $-$
& \cellcolor{beige} \meanstd{72.4}{0.4}
& \cellcolor{beige} \meanstd{8.2}{0.2}
& \cellcolor{beige} $14.6$
& \cellcolor{dusty} \meanstd{72.1}{0.9}
& \cellcolor{dusty} \meanstd{8.2}{0.1}
& $-$ \\

$N=25$
& \cellcolor{bluegray} \meanstd{84.3}{2.4}
& \cellcolor{bluegray} \meanstd{7.9}{0.2}
& $-$
& \cellcolor{beige} \meanstd{\uline{77.6}}{0.8}
& \cellcolor{beige} \meanstd{\uline{8.1}}{0.2}
& \cellcolor{beige} $\uline{6.7}$
& \cellcolor{dusty} \meanstd{78.3}{0.5}
& \cellcolor{dusty} \meanstd{\uline{8.1}}{0.1}
& $-$ \\

$N=50$
& \cellcolor{bluegray} \meanstd{80.5}{0.4}
& \cellcolor{bluegray} \meanstd{7.9}{0.1}
& $-$
& \cellcolor{beige} \meanstd{\mathbf{77.9}}{0.9}
& \cellcolor{beige} \meanstd{\mathbf{8.0}}{0.2}
& \cellcolor{beige} $\mathbf{2.6}$
& \cellcolor{dusty} \meanstd{\mathbf{79.2}}{0.3}
& \cellcolor{dusty} \meanstd{\mathbf{7.9}}{0.1}
& $-$ \\

$N=100^*$
& $-$ & $-$ & $-$
& $-$ & $-$ & $-$
& \cellcolor{bluegray} \meanstd{\uline{78.9}}{0.9}
& \cellcolor{bluegray} \meanstd{\uline{8.1}}{0.1}
& $-$ \\

\midrule

\multirow{2}{*}[-3pt]{Policy}
& \multicolumn{3}{c}{SimplerEnv scenes (3)}
& \multicolumn{3}{c}{Google scanned scenes (18)}
& \multicolumn{3}{c}{Hunyuan3D-2.1 scenes (25)} \\
\cmidrule(lr){2-4}
\cmidrule(lr){5-7}
\cmidrule(lr){8-10}
& SR (\%) [$\uparrow$] & TF (s) [$\downarrow$] & $\Delta$G
& SR (\%) [$\uparrow$] & TF (s) [$\downarrow$] & $\Delta$G
& SR (\%) [$\uparrow$] & TF (s) [$\downarrow$] & $\Delta$G \\
\midrule

$N=0 \enspace (\pi_\textrm{pre})$
& $23.7$ & $9.6$ & $-$
& $12.1$ & $\mathbf{8.1}$ & $-$
& $10.2$ & $9.0$ & $-$ \\

$N=3 \enspace (\textrm{SE})$
& \cellcolor{bluegray} $\mathbf{96.7}$
& \cellcolor{bluegray} $\mathbf{6.7}$
& $-$
& \cellcolor{beige} $35.2$
& \cellcolor{beige} $\uline{8.6}$
& \cellcolor{beige} $61.5$
& \cellcolor{beige} $38.4$
& \cellcolor{beige} $\mathbf{8.5}$
& \cellcolor{beige} $58.3$ \\

$N=1$
& \cellcolor{beige} \meanstd{36.1}{10.4}
& \cellcolor{beige} \meanstd{9.3}{1.7}
& \cellcolor{beige} $58.2$
& \cellcolor{beige} \meanstd{42.3}{4.8}
& \cellcolor{beige} \meanstd{9.3}{0.2}
& \cellcolor{beige} $52.0$
& \cellcolor{beige} \meanstd{42.7}{3.8}
& \cellcolor{beige} \meanstd{9.6}{0.3}
& \cellcolor{beige} $51.6$ \\

$N=3$
& \cellcolor{beige} \meanstd{47.4}{2.3}
& \cellcolor{beige} \meanstd{9.8}{0.6}
& \cellcolor{beige} $41.3$
& \cellcolor{beige} \meanstd{48.8}{7.3}
& \cellcolor{beige} \meanstd{9.3}{0.3}
& \cellcolor{beige} $39.9$
& \cellcolor{beige} \meanstd{50.0}{5.0}
& \cellcolor{beige} \meanstd{9.4}{0.2}
& \cellcolor{beige} $38.7$ \\

$N=10$
& \cellcolor{beige} \meanstd{54.3}{9.2}
& \cellcolor{beige} \meanstd{9.2}{0.4}
& \cellcolor{beige} $32.7$
& \cellcolor{beige} \meanstd{59.3}{2.6}
& \cellcolor{beige} \meanstd{8.8}{0.1}
& \cellcolor{beige} $27.7$
& \cellcolor{beige} \meanstd{58.2}{2.9}
& \cellcolor{beige} \meanstd{\uline{8.9}}{0.2}
& \cellcolor{beige} $28.8$ \\

$N=25$
& \cellcolor{beige} \meanstd{70.1}{5.4}
& \cellcolor{beige} \meanstd{8.4}{0.1}
& \cellcolor{beige} $14.2$
& \cellcolor{beige} \meanstd{\mathbf{62.7}}{0.6}
& \cellcolor{beige} \meanstd{8.8}{0.1}
& \cellcolor{beige} $21.6$
& \cellcolor{beige} \meanstd{61.8}{0.3}
& \cellcolor{beige} \meanstd{9.0}{0.1}
& \cellcolor{beige} $22.5$ \\

$N=50$
& \cellcolor{beige} \meanstd{68.4}{5.5}
& \cellcolor{beige} \meanstd{8.5}{0.6}
& \cellcolor{beige} $\uline{12.1}$
& \cellcolor{beige} \meanstd{\uline{62.5}}{0.4}
& \cellcolor{beige} \meanstd{8.8}{0.1}
& \cellcolor{beige} $\mathbf{18.0}$
& \cellcolor{beige} \meanstd{\mathbf{63.3}}{0.4}
& \cellcolor{beige} \meanstd{\uline{8.9}}{0.2}
& \cellcolor{beige} $\uline{17.2}$ \\

$N=100^*$
& \cellcolor{beige} \meanstd{\uline{74.3}}{4.1}
& \cellcolor{beige} \meanstd{\uline{8.3}}{0.3}
& \cellcolor{beige} $\mathbf{4.6}$
& \cellcolor{beige} \meanstd{60.7}{0.9}
& \cellcolor{beige} \meanstd{9.0}{0.1}
& \cellcolor{beige} $\uline{18.2}$
& \cellcolor{beige} \meanstd{\uline{62.5}}{1.4}
& \cellcolor{beige} \meanstd{\uline{8.9}}{0.0}
& \cellcolor{beige} $\mathbf{16.4}$ \\

\bottomrule
\end{NiceTabular}

\label{tab:sim_results}
\end{table*}

\begin{wrapfigure}{r}{0.45\textwidth}
\vspace{-14pt}
\includegraphics[width=0.45\columnwidth]{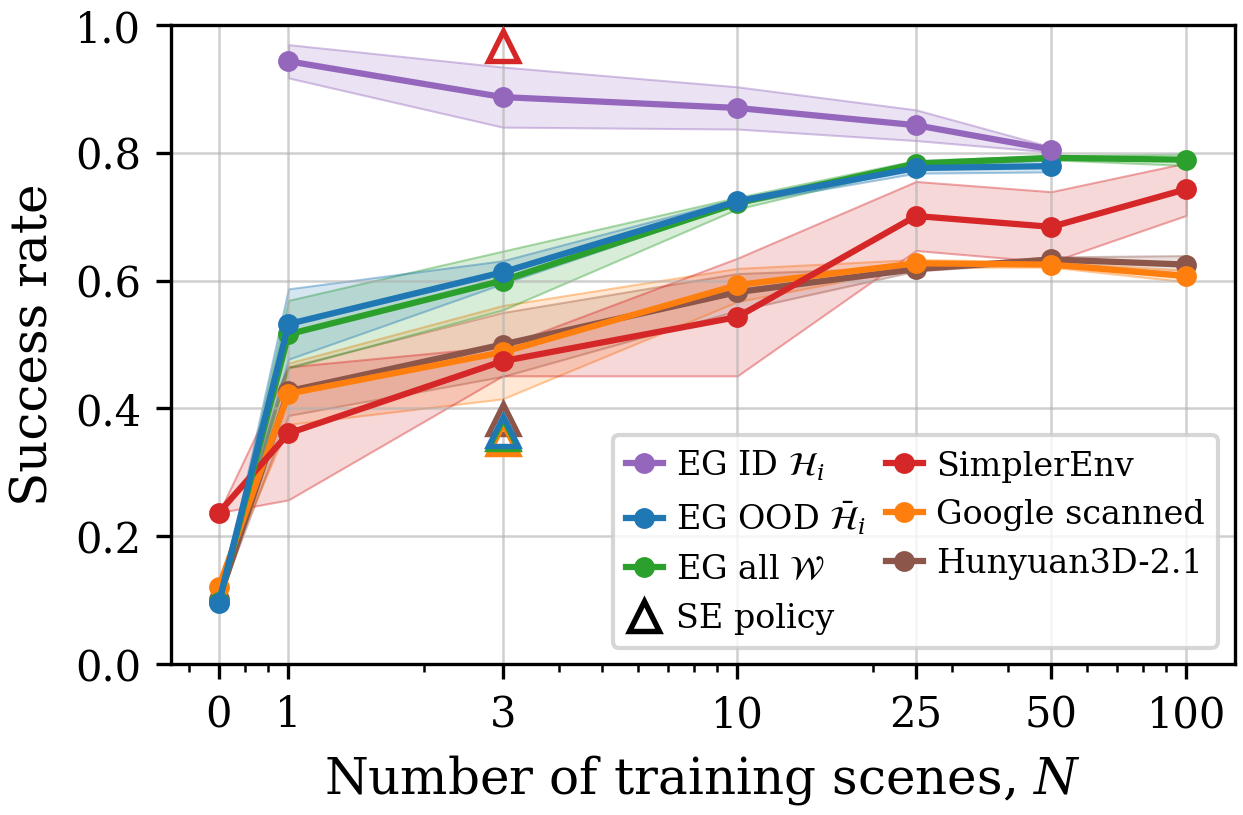}
\caption{Success rate as a function of $N$.}
\vspace{-8pt}
\label{fig:scene_scaling}
\end{wrapfigure}

\bl{Effect of number of scenes $N$.} Table~\ref{tab:sim_results} reports simulation results as a function of the number of training scenes $N$, while Fig.~\ref{fig:scene_scaling} visualizes the corresponding success-rate scaling trends.
We evaluate average success rate (SR), average time to finish (TF), and the ID--OOD generalization gap $\Delta$G across six sets of evaluation scenes.
TF is averaged over successful episodes only.
Because $\Delta$G can also decrease as ID performance falls, we interpret it jointly with absolute OOD success rather than as a standalone measure of generalization.
Results reported with standard deviation are averaged over three independent runs. For $N \leq 50$, these runs use different scene subsets $\mathcal{H}_i$, while $N=100$ uses three random seeds over the full set $\mathcal W$.
Each scene is evaluated over 100 episodes.

For policies trained on EmbodiedGen (EG) scenes, increasing $N$ produces a clear trade-off between in-distribution (ID) and out-of-distribution (OOD) performance.
As $N$ increases up to $N=50$, success on the training scenes $\mathcal H_i$ decreases under the fixed compute budget, while absolute OOD performance improves substantially.
For example, ID success decreases from 94.3\% at $N=1$ to 80.5\% at $N=50$, while EG OOD success improves from 53.2\% to 77.9\%.
Correspondingly, $\Delta$G decreases monotonically from 41.1 points at $N=1$ to 27.4 at $N=3$, 14.6 at $N=10$, 6.7 at $N=25$, and only 2.6 points at $N=50$.
Together, the increasing OOD success rates and shrinking ID--OOD gaps indicate that increasing scene diversity reduces over-specialization to the training environments.

This trend extends to OOD benchmarks spanning asset-source and generator shifts.
On SimplerEnv, success increases from 36.1\% at $N=1$ to 74.3\% at $N=100$, while $\Delta$G decreases from 58.2 to 4.6 points.
On Google Scanned Objects, success increases from 42.3\% at $N=1$ to 62.7\% at $N=25$, while $\Delta$G decreases from 52.0 to 21.6 points.
On Hunyuan3D-2.1, success increases from 42.7\% at $N=1$ to 63.3\% at $N=50$, while $\Delta$G decreases from 51.6 to 17.2 points.
As seen in Fig.~\ref{fig:scene_scaling}, the largest gains in absolute OOD success occur by approximately $N=25$--$50$, after which performance begins to saturate.
At $N=100$, success remains comparable on EG and Hunyuan3D-2.1, decreases slightly on Google Scanned Objects, and increases further on SimplerEnv to 74.3\%.
This saturation is consistent with the fixed-compute trade-off discussed above, where additional scene diversity comes at the cost of reduced per-scene coverage.

The policy trained on the three manually designed SimplerEnv scenes achieves 96.7\% success on those same scenes but transfers substantially worse to EG (36.0\%), Google Scanned Objects (35.2\%), and Hunyuan3D-2.1 (38.4\%), with corresponding $\Delta$G values of 60.7, 61.5, and 58.3 points, respectively.
In comparison, the $N=3$ EG policies achieve $\Delta$G values ranging from 27.4 to 41.3 points, highlighting the limited coverage of the three manually designed scenes compared to the broader distribution provided by EG-generated environments.
Overall, policies trained on 50--100 generated scenes achieve approximately 79\% success on $\mathcal W$, compared to 9.7\% for the imitation baseline $\pi_\textrm{pre}$, while reducing average completion time from about 10\,s to 8\,s.
Though some RL-in-real works report success rates exceeding 90\%, these results are typically obtained in narrowly defined settings~\citep{chen2025conrft, yang2024robofume}.
Our goal is not to saturate such constrained settings, but to demonstrate that generative simulation enables a scalable training distribution for RL fine-tuning of VLAs.

\begin{table*}[!t]
\renewcommand{\arraystretch}{1.1}
\caption{Sim-to-real evaluation results. 
Objects or attributes that are OOD during RL training are \mbe{highlighted}.
Attributes in parentheses were not included in the language command.
Columns are color coded as
{\setlength{\fboxsep}{1pt}\colorbox{bluegray}{success}},
{\setlength{\fboxsep}{1pt}\colorbox{beige}{failure}}, and
{\setlength{\fboxsep}{1pt}\colorbox{dusty}{time}} metrics.
}
\centering
\scriptsize
\begin{tabular*}{\textwidth}{@{\extracolsep{\fill}} l ll cccc}
\toprule

Scene & \multicolumn{2}{l}{Language Command} & \multicolumn{4}{c}{Scene Distractors} \\
\midrule
0 & \multicolumn{2}{l}{``put the banana on the bowl."} & \multicolumn{4}{l}{broccoli, strawberry, plate} \\
1 & \multicolumn{2}{l}{``put the broccoli on the mug."} & \multicolumn{4}{l}{knife, cutting board} \\
2 & \multicolumn{2}{l}{``put the mushroom on the bowl."} & \multicolumn{4}{l}{banana, broccoli, plate} \\
3 & \multicolumn{2}{l}{``put the spoon on the napkin."} & \multicolumn{4}{l}{fork, plate} \\
4 & \multicolumn{2}{l}{``put the (pink) eraser on the notebook."} & \multicolumn{4}{l}{black pen, mug} \\
5 & \multicolumn{2}{l}{``put the \mbe{(yellow) brush} on the bowl."} & \multicolumn{4}{l}{banana, \mbe{yellow} marker} \\
6 & \multicolumn{2}{l}{``put the \mbe{white} eraser on the mug."} & \multicolumn{4}{l}{pink eraser, black pen, \mbe{blue} pen} \\
7 & \multicolumn{2}{l}{``put the \mbe{red} knife on the cutting board."} & \multicolumn{4}{l}{regular knife} \\
8 & \multicolumn{2}{l}{``put the \mbe{blue} pen on the \mbe{(blue)} bowl."} & \multicolumn{4}{l}{black pen, white eraser, basket} \\
9 & \multicolumn{2}{l}{``put the \mbe{green} marker on the basket."} & \multicolumn{4}{l}{red marker, \mbe{yellow} marker, \mbe{dry erase eraser}, \mbe{cardboard box}} \\
10 & \multicolumn{2}{l}{``put the \mbe{screwdriver} on the basket."} & \multicolumn{4}{l}{knife, black marker} \\
11 & \multicolumn{2}{l}{``put the \mbe{blue} teacup on the \mbe{yellow} teacup."} & \multicolumn{4}{l}{\mbe{purple} teacup, \mbe{pink} teapot, \mbe{blue} plate, \mbe{yellow} plate} \\
\end{tabular*}

\setlength{\tabcolsep}{6.78pt}
\begin{tabular*}{\textwidth}{@{\extracolsep{\fill}} l >{\columncolor{bluegray}}c >{\columncolor{bluegray}}c >{\columncolor{beige}}c >{\columncolor{beige}}c >{\columncolor{dusty}}c  >{\columncolor{bluegray}}c >{\columncolor{bluegray}}c >{\columncolor{beige}}c >{\columncolor{beige}}c >{\columncolor{dusty}}c}
\midrule
\multirow{2}{*}[-3pt]{} & \multicolumn{5}{c}{Imitation baseline $\pi_\textrm{pre}$} & \multicolumn{5}{c}{$N=100$} \\
\cmidrule(lr){2-6}
\cmidrule(lr){7-11}
& PSR [$\uparrow$] & SR [$\uparrow$] & DFR [$\downarrow$] & SFR [$\downarrow$] & TF (s) [$\downarrow$] & PSR [$\uparrow$] & SR [$\uparrow$] & DFR [$\downarrow$] & SFR [$\downarrow$] & TF (s) [$\downarrow$] \\
\midrule
$0$ & $0.8$ & $0.6$ & $0.4$ & $0.0$ & \meanstd{10.7}{6.3} & $\mathbf{0.9}$  & $\mathbf{0.9}$ & $0.1$ & $0.0$ & \meanstd{\mathbf{7.4}}{1.5}  \\
$1$ & $0.4$ & $0.2$ & $0.8$ & $0.0$ & \meanstd{13.3}{2.0} & $\mathbf{1.0}$ & $\mathbf{0.7}$ & $0.3$ & $0.0$ & \meanstd{\mathbf{10.4}}{7.5} \\
$2$ & $0.3$ & $0.1$ & $0.9$ & $0.0$ & $11.5$ & $\mathbf{0.9}$ & $\mathbf{0.9}$ & $0.1$ & $0.0$ & \meanstd{\mathbf{8.4}}{2.5} \\
$3$ & $0.2$ & $0.1$ & $0.7$ & $0.5$ & $\mathbf{6.1}$ & $\mathbf{0.8}$ & $\mathbf{0.7}$ & $0.1$ & $0.2$ & \meanstd{12.2}{5.1} \\
$4$ & $0.4$ & $0.3$ & $0.7$ & $0.1$ & \meanstd{11.0}{4.9} & $\mathbf{1.0}$ & $\mathbf{1.0}$ & $-$ & $-$ & \meanstd{\mathbf{10.2}}{4.2} \\
$5$ & $0.6$ & $0.3$ & $0.7$ & $0.0$ & \meanstd{10.5}{1.1}  & $\mathbf{1.0}$ & $\mathbf{1.0}$ & $-$ & $-$ & \meanstd{\mathbf{9.5}}{3.9}  \\
$6$ & $0.4$ & $0.2$ & $0.8$ & $0.0$ & \meanstd{\mathbf{7.8}}{0.1} & $\mathbf{1.0}$ & $\mathbf{0.7}$ & $0.3$ & $0.0$ & \meanstd{11.1}{5.5} \\
$7$ & $0.7$ & $0.2$ & $0.2$ & $0.6$ & \meanstd{14.6}{0.6} & $\mathbf{0.8}$ & $\mathbf{0.8}$  & $0.0$  & $0.2$ & \meanstd{\mathbf{12.2}}{6.5} \\
$8$ & $0.3$ & $0.1$ & $0.9$ & $0.0$ & $29.6$ & $\mathbf{0.9}$ & $\mathbf{0.7}$ & $0.2$ & $0.1$ & \meanstd{\mathbf{12.2}}{7.4} \\
$9$ & $0.6$ & $0.3$ & $0.6$ & $0.1$ & \meanstd{9.2}{1.8} & $\mathbf{0.7}$ & $\mathbf{0.6}$ & $0.4$ & $0.0$ & \meanstd{\mathbf{8.8}}{2.0} \\
$10$ & $0.2$ & $0.0$ & $0.8$ & $0.3$ & $-$ & $\mathbf{0.6}$ & $\mathbf{0.5}$ & $0.2$ & $0.3$ & \meanstd{10.4}{5.5} \\
$11$ & $0.5$ & $0.2$ & $0.5$ & $0.6$ & \meanstd{11.6}{2.0} & $\mathbf{1.0}$ & $\mathbf{0.5}$ & $0.5$ & $0.0$ & \meanstd{\mathbf{11.0}}{2.0} \\
\midrule
All & $0.45$ & $0.217$ & $0.667$ & $0.183$ & \meanstd{11.5}{5.5} 
& $\mathbf{0.883}$ & $\mathbf{0.75}$ & $0.183$ & $0.067$ & \meanstd{\mathbf{10.2}}{5.1} \\

\bottomrule
\vspace{-15pt}
\end{tabular*}
\label{tab:sim_to_real}
\end{table*}

\bl{Sim-to-real performance.}
For real-world evaluation, we deploy both the initial imitation policy $\pi_\textrm{pre}$ and an $N=100$ checkpoint that achieves 79.8\% success on the full simulation scene set $\mathcal W$.
Table~\ref{tab:sim_to_real} reports sim-to-real evaluation results across 12 scenes and 240 real-world trials.
Each scene is evaluated using 10 trials per policy.
We report partial success rate (PSR), defined as correctly grasping and lifting the target object, overall task success rate (SR), dynamics failure rate (DFR), semantic failure rate (SFR), and time to finish (TF).
DFR measures failures caused by execution errors such as inaccurate grasp attempts or dropped objects, while SFR measures failures caused by incorrect task interpretation (e.g., interacting with the wrong object).
These failure categories are not mutually exclusive, as a single failure may involve both semantic and dynamics-related errors.
Qualitative examples of the real-world scenes are provided in Sec.~\ref{sec:sim_to_real_scenes}.

Overall performance improves substantially after RL fine-tuning.
Partial success increases from 45\% for $\pi_\textrm{pre}$ to 88.3\%, indicating a large improvement in reliable object acquisition.
Overall task success improves even more, increasing from 21.7\% to 75\%.
In addition to higher success rates, the RL policy produces more efficient behavior, reducing the average completion time from 11.5\,s to 10.2\,s.
Examining the failure breakdown reveals that most errors of the imitation baseline arise from dynamics-related issues.
Across all trials, the baseline exhibits a DFR of 66.7\%, which decreases to 18.3\% after RL fine-tuning, indicating a substantial improvement in manipulation robustness.
Semantic failures are comparatively less frequent but also decrease from 18.3\% to 6.7\%, suggesting improved task grounding and object selection.

Performance gains are consistent across nearly all scenes.
For example, scene~10 introduces a screwdriver that does not appear in the RL training distribution, yet the RL policy achieves an SR of 50\% compared to 0\% for the baseline.
Similarly, scene~11 evaluates a teacup stacking task involving unseen object instances and task composition, where the RL policy improves SR from 20\% to 50\% while maintaining perfect partial success (100\%).
Overall, these results demonstrate that policies fine-tuned in large-scale generative simulation transfer effectively to real-world manipulation.
RL fine-tuning improves both low-level execution robustness and high-level task success while maintaining generalization to previously unseen objects, attributes, and task variations.
Qualitative examples of the imitation and RL policy rollouts across different scenes are shown in Sec.~\ref{sec:real_world_rollouts}.

\section{Conclusion}
In this work, we explored the use of 3D world generative models for scaling sim-to-real reinforcement learning of vision–language–action (VLA) policies. 
By leveraging generative simulation to automatically construct diverse interactive environments, we fine-tuned a pretrained VLA across 100 unique scenes using massively parallel RL while avoiding the cost of manual scene design.
Our results show that increasing scene diversity substantially improves OOD generalization across EmbodiedGen, SimplerEnv, Google Scanned Objects, and Hunyuan3D-2.1 evaluation scenes, with the largest gains occurring by approximately $N=25$--$50$ under a fixed compute budget.
Furthermore, RL fine-tuning on the generated scenes improves simulation success rates by up to 70.1 percentage points, while sim-to-real deployment yields a 53.3-percentage-point improvement in real-world task success.
Beyond improved performance, our findings highlight the complementary roles of generative simulation and RL. 
Generative 3D world models provide a scalable source of diverse training environments, while RL enables efficient adaptation of large pretrained policies using only sparse rewards. 
Our language-driven scene designer further reduces engineering effort by automatically translating task descriptions into structured environments for simulation.
Together, these components form a practical pipeline for scaling robot foundation models while minimizing the need for additional real-world data collection or human-in-the-loop training.

\bl{Limitations and future work.} 
We note that the experiments in this work focus on pick-and-place tasks, primarily due to current limitations in the types of objects that EmbodiedGen~\citep{wang2025embodiedgengenerative3dworld, wang2026embodiedgenv2agenticsimulationready} can reliably generate.
Pick-and-place tasks also constitute a large portion of existing robot datasets, accounting for over 70\% of BridgeV2.
More broadly, our experiments do not establish that the observed scaling trends generalize across manipulation skills, VLA backbones, embodiments, or all forms of distribution shift.
In particular, we primarily vary backgrounds and object assets; shifts in task semantics, language phrasing and composition, robot embodiment, and interaction dynamics remain for future study.
Additionally, the training compute budget is fixed across $N$, causing per-scene coverage to decrease as the number of training scenes grows.
Consequently, this setup disadvantages larger $N$ policies and the observed saturation at $N=25$--$50$ may partly reflect compute limitations rather than an inherent limit to the benefits of scene diversity.
Looking forward, we aim to extend the current framework to support richer manipulation behaviors such as articulated object manipulation, tool use, and multi-stage tasks.
Scaling the distribution of automatically generated scenes and tasks will enable RL fine-tuning of VLAs on a broader range of interaction distributions, further improving generalization and robustness.
Nevertheless, we believe this work represents an important first step toward demonstrating the potential of combining generative simulation with RL to scale the fine-tuning of VLA models.

\newpage

\bibliography{references}  

\newpage

\appendix
\counterwithin{table}{section}
\counterwithin{figure}{section}
\counterwithin{equation}{section}
\renewcommand{\thetable}{\thesection.\arabic{table}}
\renewcommand{\thefigure}{\thesection.\arabic{figure}}
\renewcommand{\theequation}{\thesection.\arabic{equation}}

\section{Training Details}
\label{sec:appendix_training_details}
In this section, we list the training parameters used for all policies in Table~\ref{tab:training_parameters}.
We also summarize several practical observations encountered during training:
\begin{enumerate}[leftmargin=12pt, itemsep=2pt, topsep=2pt]
    \item Several training strategies were evaluated, including:
    \begin{enumerate}[itemsep=2pt, topsep=2pt]
        \item Freezing the VLM and LoRA fine-tuning the action head.
        \item Freezing the VLM and fully fine-tuning the action head.
        \item Freezing the SigLip vision encoder, LoRA fine-tuning Gemma, and LoRA fine-tuning the action head.
        \item LoRA fine-tuning the VLM and LoRA fine-tuning the action head.
        \item LoRA fine-tuning the VLM and fully fine-tuning the action head.
    \end{enumerate}
    Overall, we found that freezing the VLM (a, b) significantly lowers the training ceiling and often leads to model collapse during prolonged training, where task success rates initially peak, then saturate, and eventually decline toward zero.
    As a result, updating the VLM is crucial for stable learning.
    Fully fine-tuning the action head consistently outperformed the corresponding configuration that LoRA fine-tunes it.
    Freezing the SigLip vision encoder reduced success rates monotonically by a few percent compared to LoRA fine-tuning it.
    Since LoRA fine-tuning the entire VLM did not noticeably affect sim-to-real performance, we adopt option (e) in this work.
    
    \item Adding the scaling exponent $s$ to the importance ratio (Eq.~\ref{eq:importance_ratio}) is crucial for stabilizing training and preventing immediate collapse.
    Without it, the computed log probabilities can spike to very large values, causing many updates to fall near the clipping threshold and resulting in the majority of gradients being clipped.
    
\end{enumerate}

\begin{table}[h]
\renewcommand{\arraystretch}{1.1}
\caption{Training parameters.}
\centering
\footnotesize
\begin{tabular}{lr|lr}
\toprule
Parameter & Value & Parameter & Value \\
\midrule
number of environments & $192$ & action chunk size $C$ & $4$ \\
batch size & $19200$ & number of integration steps $K$ & $1$ \\
mini batch size & $1920$ & learning rate & $2\mathrm{e-}5$ \\
episode length & $25$ & gradient global norm clip & $0.5$ \\
discount factor $\gamma$ & $0.99$ & clipping ratio $\epsilon$ & $0.2$ \\
UTD ratio & $1$ & importance ratio scale $s$ & $0.2$ \\
LoRA rank & $32$ & $\sigma_\phi$ log min & $-2.5$ \\
control frequency (Hz) & $5$ & $\sigma_\phi$ log max & $-2.0$ \\
\bottomrule
\end{tabular}
\label{tab:training_parameters}
\end{table}

\bl{Domain randomization.} Domain randomization ranges can be seen in Table~\ref{tab:domain_randomization}.
For lighting, we use the same lighting setup as the one in SimplerEnv's Bridge setting~\citep{li2024evaluatingrealworldrobotmanipulation}.
For camera position randomization, we first record the intersection point between the camera’s principal ray and the table in the default configuration.
We then compute a new camera orientation such that the principal ray of the perturbed camera re-aligns with this original intersection point.
We found that control-delay randomization was unnecessary due to the use of action chunks ($C=4$), which allow the majority of actions to be executed with minimal latency.
\begin{table}[h]
\renewcommand{\arraystretch}{1.1}
\caption{Domain randomization ranges.}
\centering
\footnotesize
\begin{tabular}{ll|ll}
\toprule
Parameter & Value & Parameter & Value \\
\midrule
object x-position (m) & $[0.2, 0.4]$ & robot z-height perturb (m) & $[0, 0.05]$ \\
object y-position (m) & $[-0.15, 0.15]$ & robot joint pos perturb (rad) & $[-0.1, 0.1]^6$ \\
object yaw orientation (rad) & $[0, 2\pi]$ & camera xyz-position (m) & $[-0.05, 0.05]^3$ \\
ambient light RGB color & $[0, 0.6]^3$ & directional light brightness & $[0.5, 1.5]$ \\
\bottomrule
\end{tabular}
\label{tab:domain_randomization}
\end{table}

\bl{Rewards.}
We employ a sparse success reward, enabled by the strong pre-training of the VLA during the imitation phase.
An episode is considered successful if
\begin{align}
\text{success} =
\text{contact}(A,B) \;\land\;
\neg\,\text{contact}(A,\text{table}) \;\land\;
\neg\,\text{contact}(A, \text{robot}),
\label{eq:reward}
\end{align}
where $A$ denotes the manipulated object and $B$ denotes the target object onto which $A$ is placed.

\section{Does Multimodality Matter? Effect of $K$}
\label{sec:k_ablation}
In this work, we set $K=1$, effectively converting the flow-matching policy $\pi_\textrm{pre}$ into a single-step Gaussian policy $\pi_\theta$ during RL fine-tuning.
While prior work such as Diffusion Steering~\citep{wagenmaker2025steeringdiffusionpolicylatent} and Flow Policy Optimization (FPO)~\citep{mcallister2026flow} demonstrate that flow-matching models can be fine-tuned without collapsing into a Gaussian policy, we found that: 
(1) PPOFlow led to significantly more stable and repeatable training, and 
(2) single-step ($K=1$) RL fine-tuning exhibited no observable degradation in policy performance compared to maintaining multimodality ($K>1$).

We validate the latter through an ablation over $K \in \{1,2,4\}$ on the entire scene--task set $\mathcal W$, where larger $K$ enables greater expressivity by chaining multiple Gaussian transitions, with $K=1$ corresponding to a unimodal policy.
From the left side of Table~\ref{tab:multi_modality}, performance remains comparable across $K \in \{1,2,4\}$, with success rates differing by only approximately 2--2.5 percentage points.
Since these results are based on a single run per setting, we do not interpret the small differences as evidence that $K=1$ is inherently superior to $K>1$.
Rather, the results show that $K=1$ preserves comparable task performance while substantially reducing computational cost.
Our findings align with recent work by \citet{pan2026much}, which argues that multimodality is not the primary driver of diffusion-based policy performance in robot manipulation.
We hypothesize that multimodality is most beneficial during imitation learning when the demonstration distribution itself is multimodal~\citep{dppo2024}. 
In contrast, under reward-guided optimization, unimodal Gaussian policies appear sufficient for effective policy improvement.

Beyond task performance, reducing $K$ substantially improves computational efficiency.
As shown in Table~\ref{tab:multi_modality}, $K=1$ achieves a 1.17$\times$ speedup in backward pass time relative to $K=2$ and a 1.45$\times$ speedup relative to $K=4$.
Inference gains are even more pronounced.
Compared to the original imitation setting of $K=10$, $K=1$ yields a 2.72$\times$ speedup in inference latency under a regular forward pass and a 2.36$\times$ speedup when using \texttt{torch.compile}.
These improvements stem from eliminating iterative denoising steps, effectively reducing the policy to a single forward pass, similar in spirit to consistency policies~\citep{chen2024boostingcontinuouscontrolconsistency, prasad2024consistencypolicyacceleratedvisuomotor, chen2025conrft}.
Overall, $K=1$ maintains comparable task performance while substantially reducing both training and inference cost.

\begin{table}[h]
\renewcommand{\arraystretch}{1.1}
\footnotesize
\caption{Effect of $K$ on RL fine-tuning and inference.
}
\centering
\begin{tabular}{l ccccc}
\toprule
\multirow{2}{*}[-3pt]{$K$} & \multirow{2}{*}[-3pt]{RL SR (\%) [$\uparrow$]} & \multirow{2}{*}[-3pt]{RL TF (s) [$\downarrow$]} & \multirow{2}{*}[-3pt]{Backward Time (s) [$\downarrow$]} & \multicolumn{2}{c}{Inference Latency (s) [$\downarrow$]} \\ 
\cmidrule(lr){5-6}
& & & & Reg. & \texttt{torch.compile} \\
\midrule
$K=10$ & N/A & N/A & N/A & $0.267$ & $0.172$ \\
$K=4$ & $77.29$ & $7.76$ & $108.80$ & $0.153$ & $0.107$ \\
$K=2$ & $77.23$ & $8.36$ & $87.55$ & $0.120$ & $0.086$ \\
$K=1$ & $79.75$ & $7.94$ & $74.74$ & $0.098$ & $0.073$ \\
\bottomrule
\end{tabular}
\label{tab:multi_modality}
\end{table}

\clearpage

\section{Generative Simulation Metrics}
\label{sec:scene_designer}

\begin{figure}[t]
\centering
\includegraphics[width=\columnwidth]{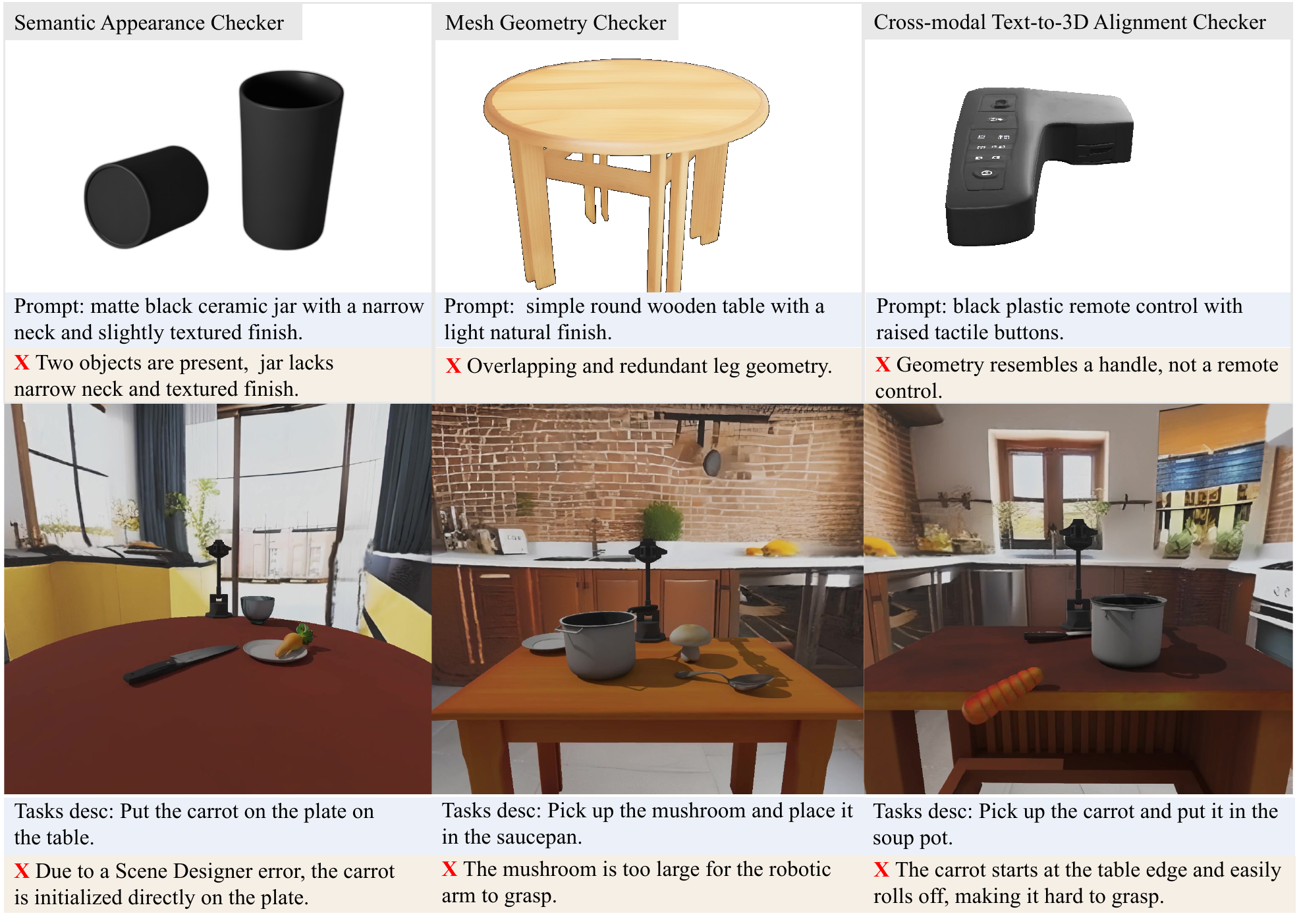}
\caption{
\textbf{Representative failure cases in the generative simulation pipeline.}
\textit{Top:} failures detected by the automated QA modules, including semantic appearance mismatch, mesh defects, and text-to-3D semantic drift.
\textit{Bottom:} residual failures identified by manual inspection after all automated checks pass, including incorrect initialization, scale mismatch, and unstable object placement.
}
\label{fig:embodiedgen_failcase}
\end{figure}

Table~\ref{tab:scene_designer} reports the profiling statistics of the generative simulation pipeline illustrated in Fig.~\ref{fig:embodiedgen_layout}. Across 100 generated environments, the system produces 516 unique object assets, averaging 5.16 interactive objects per scene. Because the asset generation pipeline proceeds from text to a background-removed foreground image and then to a 3D asset, we deploy a GPT-4o-based automated quality assurance (QA) loop at multiple stages. The \textit{Semantic Appearance checker} serves as an early filter on the intermediate foreground image, verifying whether it matches the target object category and key visual attributes. If this check fails, the system immediately resamples the text-to-image seed and retries, preventing semantically incorrect intermediate outputs from propagating into the downstream image-to-3D stage. The \textit{Mesh Geometry checker} then evaluates whether the generated 3D mesh is complete and free of major geometric defects. Finally, the \textit{Cross-modal Text-to-3D Alignment checker} assesses whether the final 3D asset remains semantically consistent with the original text description, thereby capturing semantic drift introduced during 3D generation. Combined with this GPT-4o-based QA-driven rejection-and-retry mechanism, assets require only 1.37 generation attempts on average to satisfy all constraints. Manual inspection shows that 85\% of the generated environments are directly usable for end-to-end reinforcement learning without human intervention. The remaining failures are mainly due to scale mismatches or imperfect initial object placements (representative failure case see Fig.~\ref{fig:embodiedgen_failcase}), which are generally minor and can be rectified with minimal manual adjustment. Under fully online sequential generation on a single NVIDIA RTX 4090 GPU, the pipeline requires $46.8 \pm 5.0$ minutes per scene. When reusable interactive assets are drawn from a pre-built asset library, the scene generation time is reduced to approximately 2 minutes per environment.

\begin{table}[h]
\centering
\small
\caption{\textbf{Profiling statistics of the generative simulation pipeline (Fig.~\ref{fig:embodiedgen_layout}).}
Results are aggregated over 100 generated scenes under sequential execution on a single NVIDIA RTX 4090 GPU. QA pass rates denote stage-wise single-attempt pass rates, and average attempts are measured per accepted object asset.
}
\label{tab:scene_designer}
\begin{tabular}{@{}llc@{}}
\toprule
\textbf{Category} & \textbf{Metric} & \textbf{Value} \\
\midrule
\textbf{Generation Scale}
  & Total generated environments              & 100 \\
  & Total unique 3D object assets             & 516 \\
  & Avg. interactive assets per scene         & 5.16 \\
  & Total background assets                   & 100 \\
\midrule
\textbf{Generation Efficiency}
  & Total time per scene                      & $46.8 \pm 5.0$ min \\
  & \hspace{3mm} $\llcorner$ Time per background asset  & $25.0 \pm 3.2$ min \\
  & \hspace{3mm} $\llcorner$ Time per object asset      & $3.9 \pm 1.6$ min \\
\midrule
\textbf{Automated QA Pass Rates}
  & Semantic Appearance                       & 83.3\% \\
  & Mesh Geometry                             & 75.2\% \\
  & Cross-modal Text-to-3D Alignment          & 91.9\% \\
  & \textit{Average attempts per valid asset} & 1.37 \\
\midrule
\textbf{Manual Inspection}
  & Final environment acceptance rate         & 85\% \\
\bottomrule
\end{tabular}
\end{table}

\bl{Auto-scaling of objects.}
Since the WidowX 250S manipulator (Fig.~\ref{fig:robot_fig}) used in this work has a narrow gripper width of at most 74\,mm, many generated objects are too large to be grasped.
To address this, we automatically scale down oversized objects (based on their mesh bounding boxes) to ensure they are graspable.
We also reduce mesh resolution to simplify contact computation.

\newpage

\section{Foundation Model Candidates}

In addition to $\pi_0$, we also considered OpenVLA~\citep{kim2025openvla}, but observed a significant real-to-sim performance gap\footnote{See \url{https://github.com/openvla/openvla/issues/7\#issuecomment-2330572696} for analysis by the authors on the OpenVLA real-to-sim gap.}.
Beyond this gap, the larger size of OpenVLA (7B parameters) compared to $\pi_0$ (3B) made the latter a more practical choice.
We also evaluated SpatialVLA~\citep{qu2025spatialvlaexploringspatialrepresentations}.
Although it exhibited strong real-to-sim transfer, we found its inference latency to be substantially higher than that of $\pi_0$.
Overall, we selected $\pi_0$ for its smaller model size (3B), strong real-to-sim transfer, and low inference latency.

We note that, rather than using the $\pi_0$ model released by Physical Intelligence, we adopt a model from a third-party repository (\url{https://github.com/allenzren/open-pi-zero}), as it provides a checkpoint pretrained on BridgeV2 (denoted $\pi_\textrm{pre}$).
In parallel with this paper's experiments, we also fine-tuned the official $\pi_0$ and $\pi_{0.5}$~\citep{black2025pi0.5} models from Physical Intelligence on BridgeV2.
We found that while the official $\pi_0$ achieved comparable task success to $\pi_\textrm{pre}$, the newer $\pi_{0.5}$ performed worse than both across multiple random seeds.
Thus, we use $\pi_\textrm{pre}$ throughout this work, although we expect RL fine-tuning to yield substantial gains regardless of the initial imitation policy, provided that the starting success rate is sufficiently high.

\newpage

\section{EmbodiedGen Scenes}
\label{sec:embodied_gen_scenes}

\begin{figure}[H]
\centering
\includegraphics[width=0.95\columnwidth]{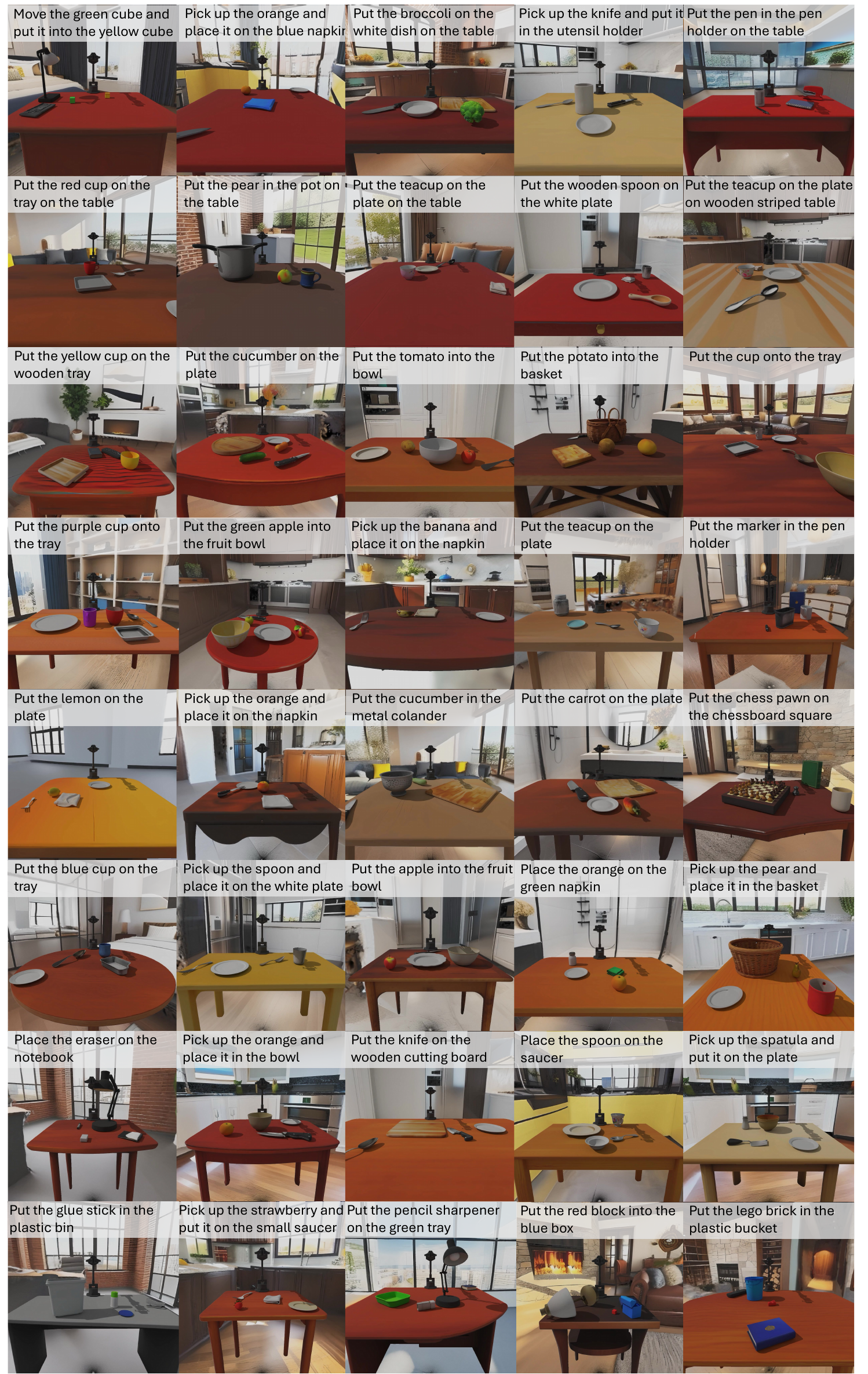}
\caption{
Generated example simulation scenes for RL fine-tuning. Zoom in for details.
}
\end{figure}

\newpage

\begin{table}[H]
\renewcommand{\arraystretch}{1.1}
\fontsize{6.5pt}{7.3pt}\selectfont
\caption{EmbodiedGen generated scenes 0--99.}
\centering

\begin{tabular}{l l l | l l l}
\toprule
Scene & Language Command & Distractors
& Scene & Language Command & Distractors \\
\midrule

$0$ & green cube $\rightarrow$ yellow cube & keyboard
& $50$ & white mouse $\rightarrow$ mouse pad & black mouse, keyboard \\

$1$ & carrot $\rightarrow$ plate & knife, bowl
& $51$ & water bottle $\rightarrow$ tray & mug, plate \\

$2$ & ceramic teapot $\rightarrow$ tray & mug, plate
& $52$ & apple $\rightarrow$ fruit plate & glass cup \\

$3$ & orange $\rightarrow$ blue napkin & plate
& $53$ & orange $\rightarrow$ round plate & fork, square plate \\

$4$ & banana $\rightarrow$ red napkin & mug, plate
& $54$ & spoon $\rightarrow$ plate & glass, fork, napkin \\

$5$ & broccoli $\rightarrow$ white dish & cutting board
& $55$ & cup $\rightarrow$ tray & spoon, plate, bowl \\

$6$ & tomato $\rightarrow$ saucepan & --
& $56$ & cup $\rightarrow$ tray & napkin, plate \\

$7$ & cucumber $\rightarrow$ metal colander & knife, cutting board
& $57$ & banana $\rightarrow$ plate & fork, orange, napkin \\

$8$ & knife $\rightarrow$ round container & spoon, plate
& $58$ & potato $\rightarrow$ basket & knife, onion, cutting board \\

$9$ & orange $\rightarrow$ blue napkin & spoon, plate
& $59$ & tomato $\rightarrow$ bowl & spoon, potato, plate \\

$10$ & red cup $\rightarrow$ tray & fork, plate
& $60$ & black pen $\rightarrow$ round container & stapler, red pencil, notebook \\

$11$ & spoon $\rightarrow$ black jar & plate
& $61$ & cucumber $\rightarrow$ plate & knife, carrot, cutting board \\

$12$ & pen $\rightarrow$ round container & book
& $62$ & blue cup $\rightarrow$ tray & spoon, plate \\

$13$ & pear $\rightarrow$ pot & mug, plate
& $63$ & spoon $\rightarrow$ white plate & fork, glass \\

$14$ & apple $\rightarrow$ plate & book
& $64$ & apple $\rightarrow$ fruit bowl & knife, plate \\

$15$ & wooden spoon $\rightarrow$ white plate & glass cup, napkin
& $65$ & orange $\rightarrow$ green napkin & salt shaker, plate \\

$16$ & teacup $\rightarrow$ plate & spoon, napkin
& $66$ & banana $\rightarrow$ plate & fork, glass \\

$17$ & orange $\rightarrow$ bowl & spoon, plate
& $67$ & lemon $\rightarrow$ metal bowl & knife, plate \\

$18$ & marker $\rightarrow$ round container & coffee mug, notebook
& $68$ & tomato $\rightarrow$ white plate & fork, napkin \\

$19$ & yellow cup $\rightarrow$ wooden tray & remote control, book
& $69$ & pear $\rightarrow$ basket & mug, plate \\

$20$ & green apple $\rightarrow$ basket & knife, plate
& $70$ & red marker $\rightarrow$ pen holder & mug, notebook \\

$21$ & white ping-pong ball $\rightarrow$ blue cup & remote control, book
& $71$ & eraser $\rightarrow$ notebook & pen \\

$22$ & cucumber $\rightarrow$ basket & knife, cutting board
& $72$ & green apple $\rightarrow$ tray & fork, plate \\

$23$ & carrot $\rightarrow$ plate & knife, cutting board
& $73$ & teacup $\rightarrow$ saucer & spoon, napkin \\

$24$ & mushroom $\rightarrow$ pot & knife, cutting board
& $74$ & knife $\rightarrow$ cutting board & spoon, plate \\

$25$ & ceramic teapot $\rightarrow$ tray & mug, plate
& $75$ & banana $\rightarrow$ basket & mug, plate \\

$26$ & cucumber $\rightarrow$ metal colander & knife, cutting board
& $76$ & lemon $\rightarrow$ napkin & fork, plate \\

$27$ & knife $\rightarrow$ utensil holder & salt shaker, plate
& $77$ & mushroom $\rightarrow$ saucepan & spoon, plate \\

$28$ & orange $\rightarrow$ napkin & knife, plate
& $78$ & potato $\rightarrow$ plate & fork, glass \\

$29$ & lemon $\rightarrow$ plate & fork, napkin
& $79$ & yellow cup $\rightarrow$ basket & spoon, plate \\

$30$ & orange $\rightarrow$ bowl & knife, cutting board
& $80$ & apple $\rightarrow$ cutting board & knife, plate \\

$31$ & tennis ball $\rightarrow$ gray basket & coffee mug, book
& $81$ & spoon $\rightarrow$ saucer & cup, plate \\

$32$ & marker $\rightarrow$ round container & coffee mug, notebook
& $82$ & spatula $\rightarrow$ plate & salt shaker, bowl \\

$33$ & yellow cup $\rightarrow$ wooden tray & fork, napkin
& $83$ & red block $\rightarrow$ blue box & lamp, book \\

$34$ & green eraser $\rightarrow$ red box & pen, notebook
& $84$ & banana $\rightarrow$ bowl & spoon, plate \\

$35$ & fork $\rightarrow$ metal tray & napkin, plate
& $85$ & pear $\rightarrow$ napkin & spoon, plate \\

$36$ & apple $\rightarrow$ basket & knife, plate
& $86$ & red apple $\rightarrow$ wicker basket & mug, plate \\

$37$ & spoon $\rightarrow$ black jar & napkin, plate
& $87$ & green lime $\rightarrow$ white bowl & knife, plate \\

$38$ & pen $\rightarrow$ pen holder & notebook
& $88$ & yellow banana $\rightarrow$ cutting board & knife, bowl \\

$39$ & pear $\rightarrow$ pot & knife, plate
& $89$ & strawberry $\rightarrow$ small saucer & fork, napkin \\

$40$ & teacup $\rightarrow$ plate & spoon
& $90$ & carrot $\rightarrow$ soup pot & knife, cutting board \\

$41$ & banana $\rightarrow$ napkin & fork, plate
& $91$ & soup spoon $\rightarrow$ empty bowl & plate, napkin \\

$42$ & wooden spoon $\rightarrow$ white plate & salt shaker, napkin
& $92$ & pepper grinder $\rightarrow$ metal tray & salt shaker, napkin \\

$43$ & marker $\rightarrow$ pen holder & coffee mug, notebook
& $93$ & red marker $\rightarrow$ white mug & mouse, keyboard \\

$44$ & green apple $\rightarrow$ basket & salt shaker, plate
& $94$ & eraser $\rightarrow$ open notebook & pen \\

$45$ & apple $\rightarrow$ basket & fork, plate
& $95$ & glue stick $\rightarrow$ plastic bin & stapler, mouse pad \\

$46$ & red cup $\rightarrow$ tray & salt shaker, napkin
& $96$ & pencil sharpener $\rightarrow$ green tray & book \\

$47$ & cucumber $\rightarrow$ basket & knife, cutting board
& $97$ & lego brick $\rightarrow$ plastic bucket & remote control, book \\

$48$ & green apple $\rightarrow$ fruit bowl & red apple, knife, plate
& $98$ & chess pawn $\rightarrow$ chessboard & mug, book \\

$49$ & purple cup $\rightarrow$ tray & spoon, red cup, plate
& $99$ & sponge $\rightarrow$ sink basin & towel \\

\bottomrule
\end{tabular}

\label{tab:generated_tasks_and_scenes}
\end{table}

\newpage

\begin{figure}[H]
\includegraphics[width=\columnwidth]{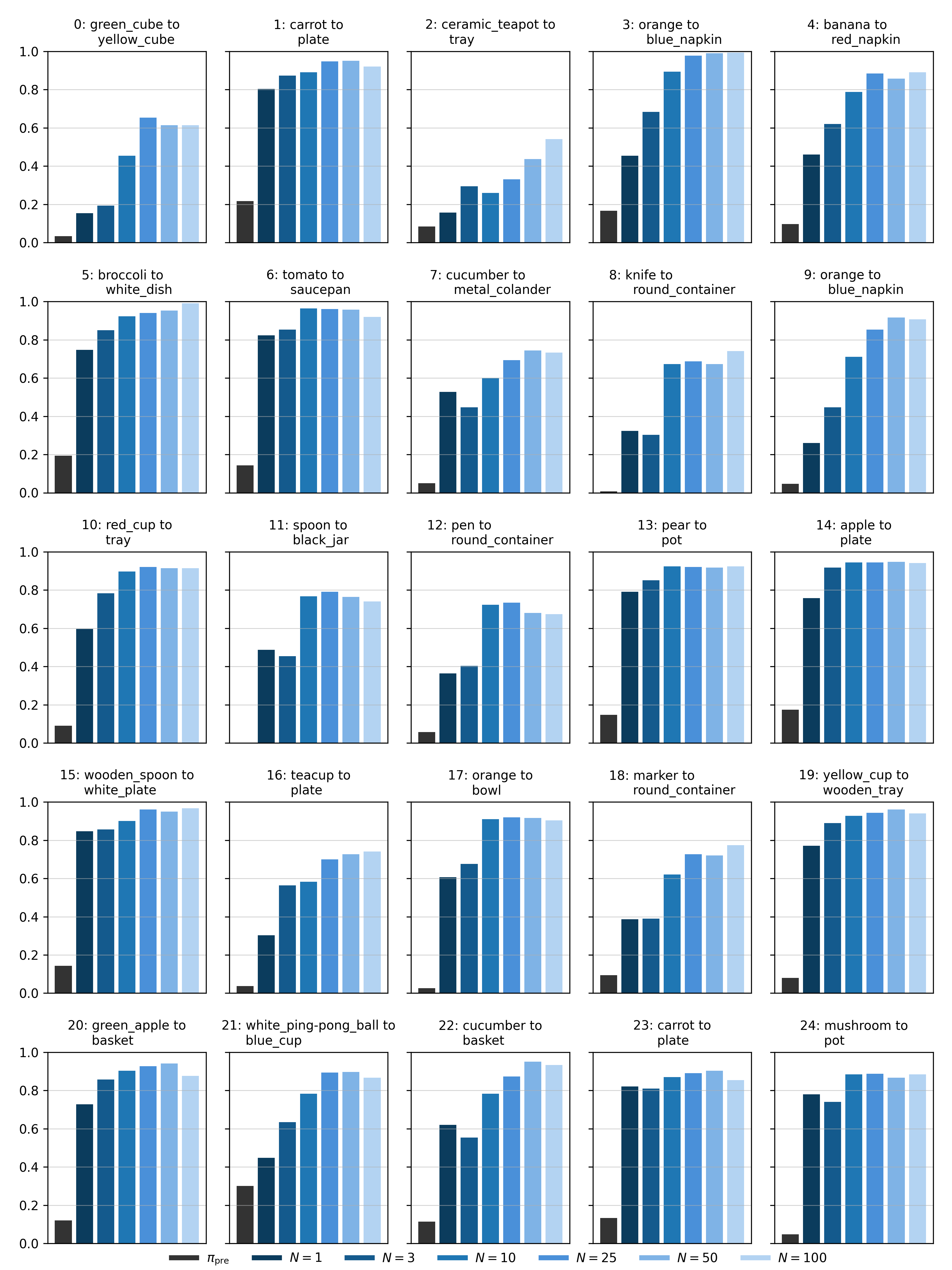}
\caption{
Success rates for EG scenes 0-24.
}
\label{fig:sr_0_24}
\end{figure}

\newpage

\begin{figure}[H]
\includegraphics[width=\columnwidth]{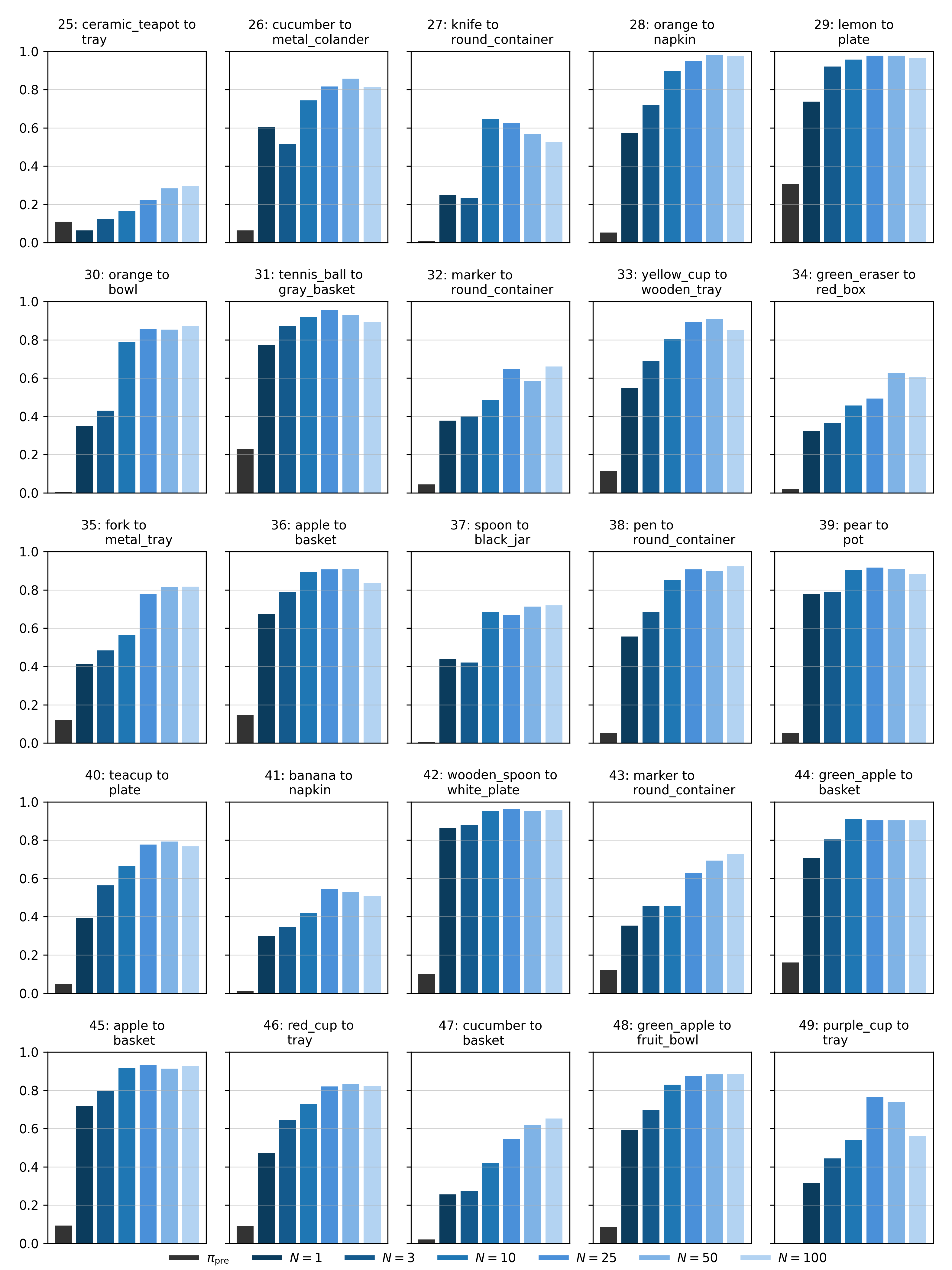}
\caption{
Success rates for EG scenes 25-49.
}
\label{fig:sr_25_49}
\end{figure}

\newpage

\begin{figure}[H]
\includegraphics[width=\columnwidth]{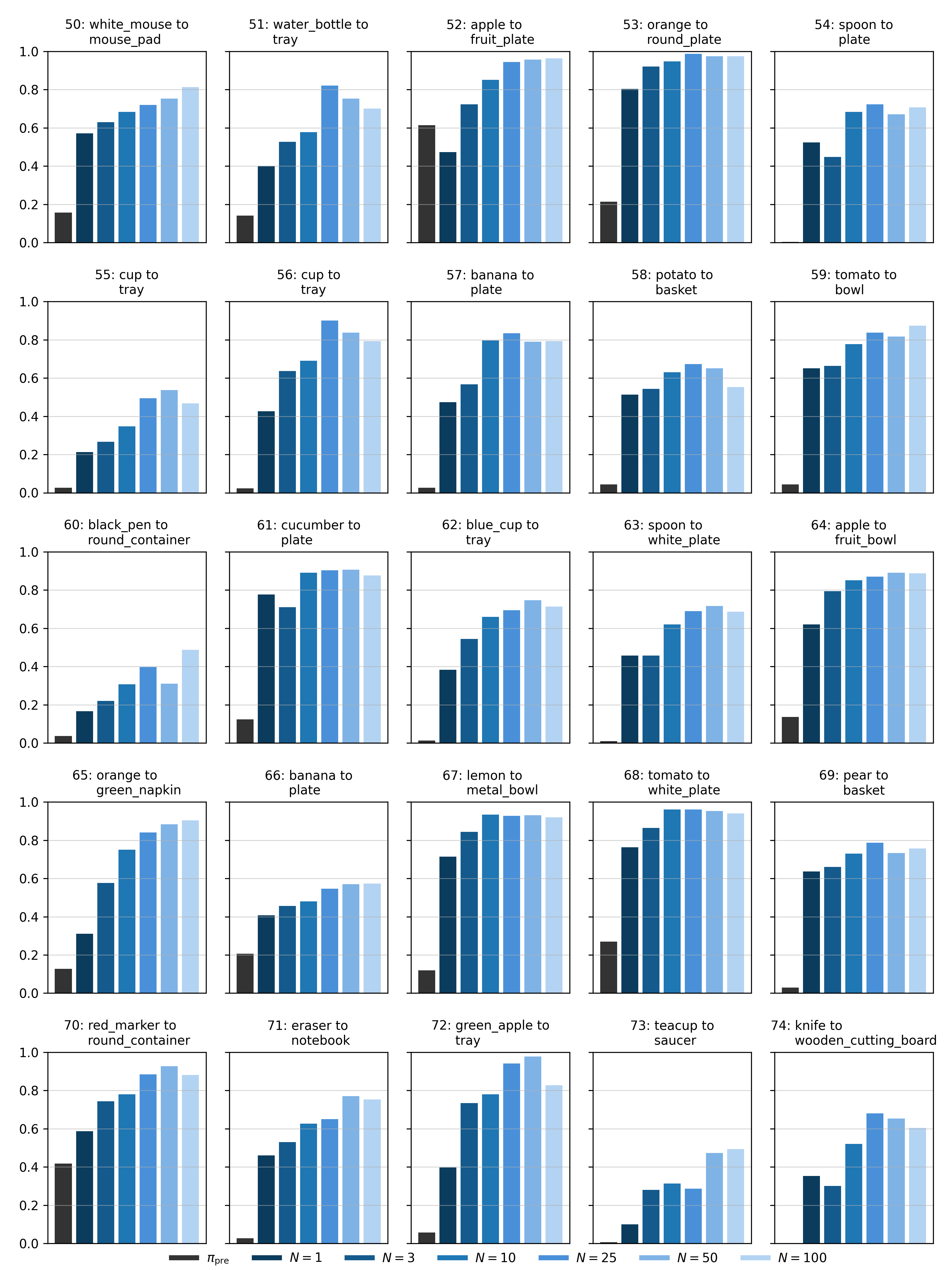}
\caption{
Success rates for EG scenes 50-74.
}
\label{fig:sr_50_74}
\end{figure}

\newpage

\begin{figure}[H]
\includegraphics[width=\columnwidth]{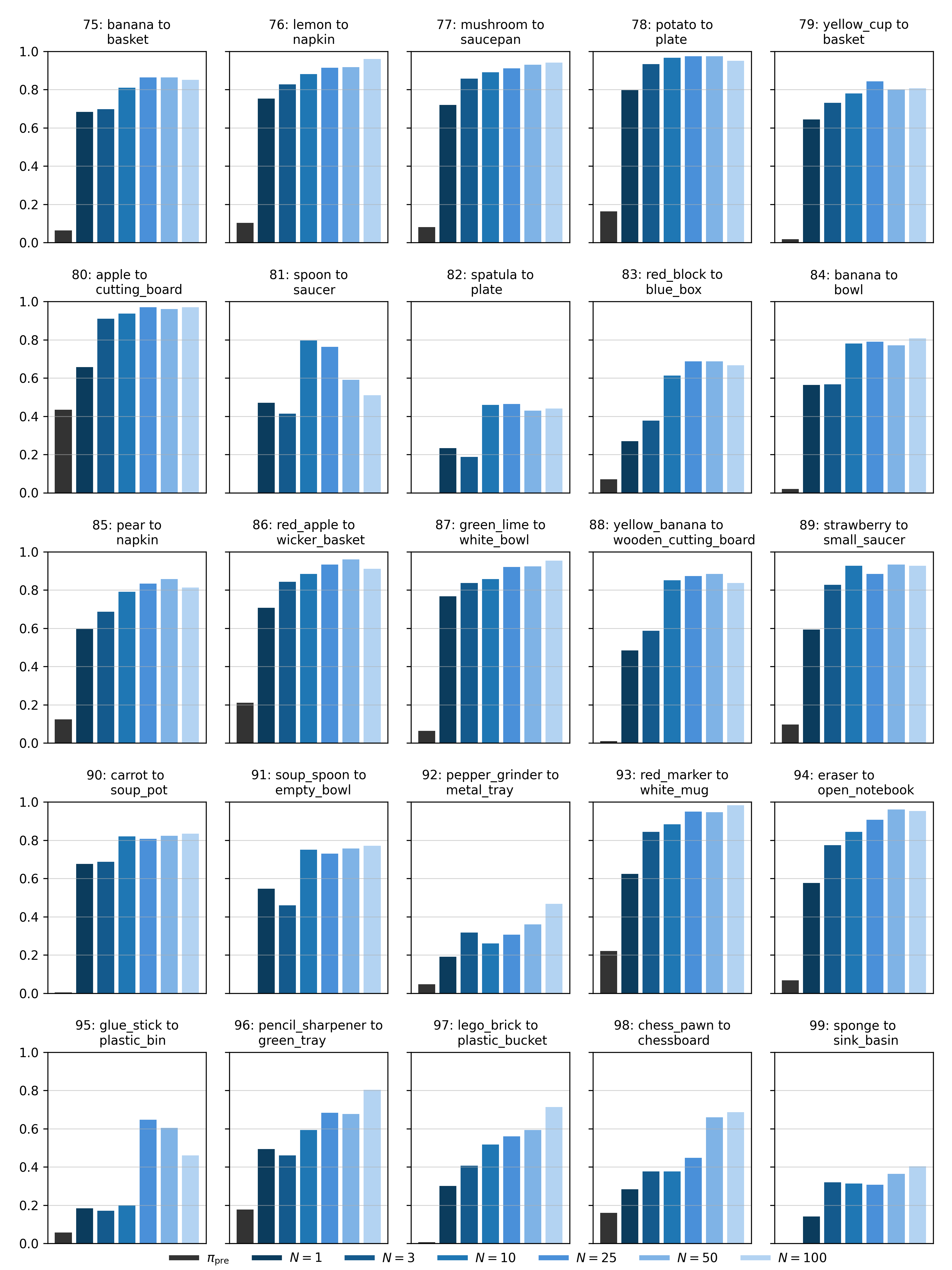}
\caption{
Success rates for EG scenes 75-99.
}
\label{fig:sr_75_99}
\end{figure}

\newpage

\section{SimplerEnv Scenes}
\label{sec:simpler_env_scenes}

\begin{figure}[h]
\centering
\includegraphics[width=\columnwidth]{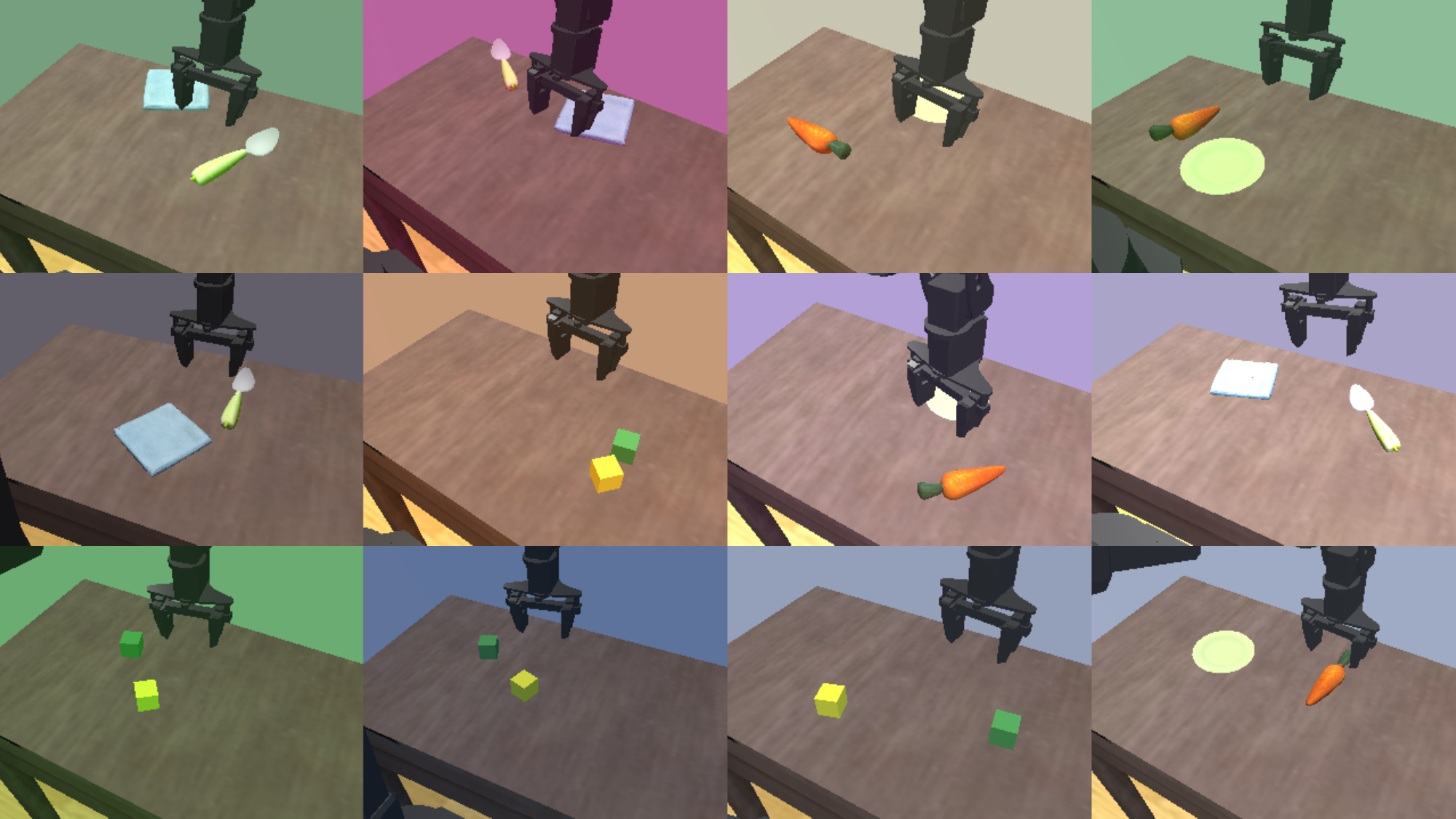}
\caption{
SimplerEnv scenes with domain randomization used for training and evaluation.
}
\label{fig:simpler_env_scenes}
\end{figure}

We use the three manually designed tabletop scenes from SimplerEnv~\citep{li2024evaluatingrealworldrobotmanipulation} both to train an $N=3$ baseline policy and to OOD evaluate EmbodiedGen-trained policies.
To be able to use the same domain randomization techniques as those in Table~\ref{tab:domain_randomization}, we remove the static png background so that camera pose and lighting randomization can properly take effect (Fig.~\ref{fig:simpler_env_scenes}).

\newpage

\section{Google Scanned Objects Scenes}
\label{sec:scanned_scenes}

\begin{figure}[H]
\centering
\includegraphics[width=\columnwidth]{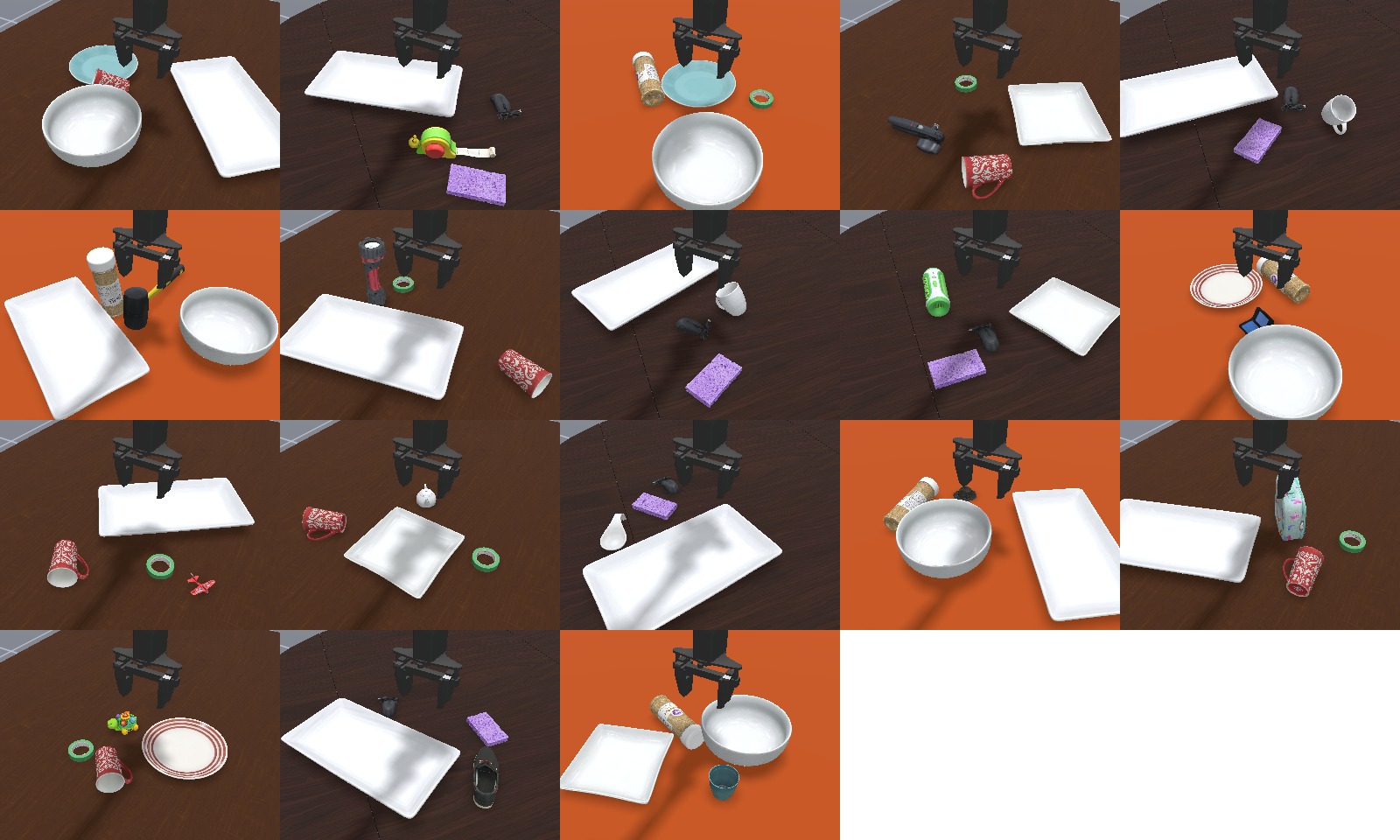}
\caption{
Google Scanned Objects scenes shown with domain randomization disabled. Evaluation was performed with domain randomization enabled.
}
\end{figure}

\begin{table}[H]
\renewcommand{\arraystretch}{1.1}
\footnotesize
\caption{Google scanned scenes 0--17.}
\centering
\begin{tabular}{l l l}
\toprule
Scene & Language Command & Distractors \\
\midrule
$0$ & green measuring tape $\rightarrow$ white tray & purple sponge, black mouse \\
$1$ & green tape $\rightarrow$ plate & gold sprinkles bottle, cereal bowl \\
$2$ & can opener $\rightarrow$ plate & coffee mug, green tape roll \\
$3$ & purple sponge $\rightarrow$ white tray & black mouse, white textured mug \\
$4$ & hammer $\rightarrow$ white tray & gold sprinkles bottle, cereal bowl \\
$5$ & flashlight $\rightarrow$ white tray & coffee mug, green tape roll \\
$6$ & black mouse $\rightarrow$ white tray & purple sponge, white textured mug \\
$7$ & green speaker $\rightarrow$ plate & purple sponge, black mouse \\
$8$ & white teapot $\rightarrow$ plate & coffee mug, green tape roll \\
$9$ & white spoon rest $\rightarrow$ white tray & purple sponge, black mouse \\
$10$ & light blue lunch box $\rightarrow$ white tray & coffee mug, green tape roll \\
$11$ & green turtle toy $\rightarrow$ plate & coffee mug, green tape roll \\
$12$ & shoe $\rightarrow$ white tray & purple sponge, black mouse \\
$13$ & red butter dish $\rightarrow$ round plate & porcelain tray, windup dog \\
$14$ & yellow school bus $\rightarrow$ plate & gold sprinkles bottle, cereal bowl \\
$15$ & dark gray rhino $\rightarrow$ white tray & windup dog, pink gum box \\
$16$ & yellow candy bar $\rightarrow$ white tray & coffee mug, green tape roll \\
$17$ & screwdriver $\rightarrow$ white tray & purple sponge, black mouse \\
\bottomrule
\end{tabular}
\label{tab:google_scanned_tasks_and_scenes}
\end{table}

\newpage

\section{Hunyuan3D-2.1 Scenes}
\label{sec:third_party_scenes}

\begin{figure}[H]
\centering
\includegraphics[width=\columnwidth]{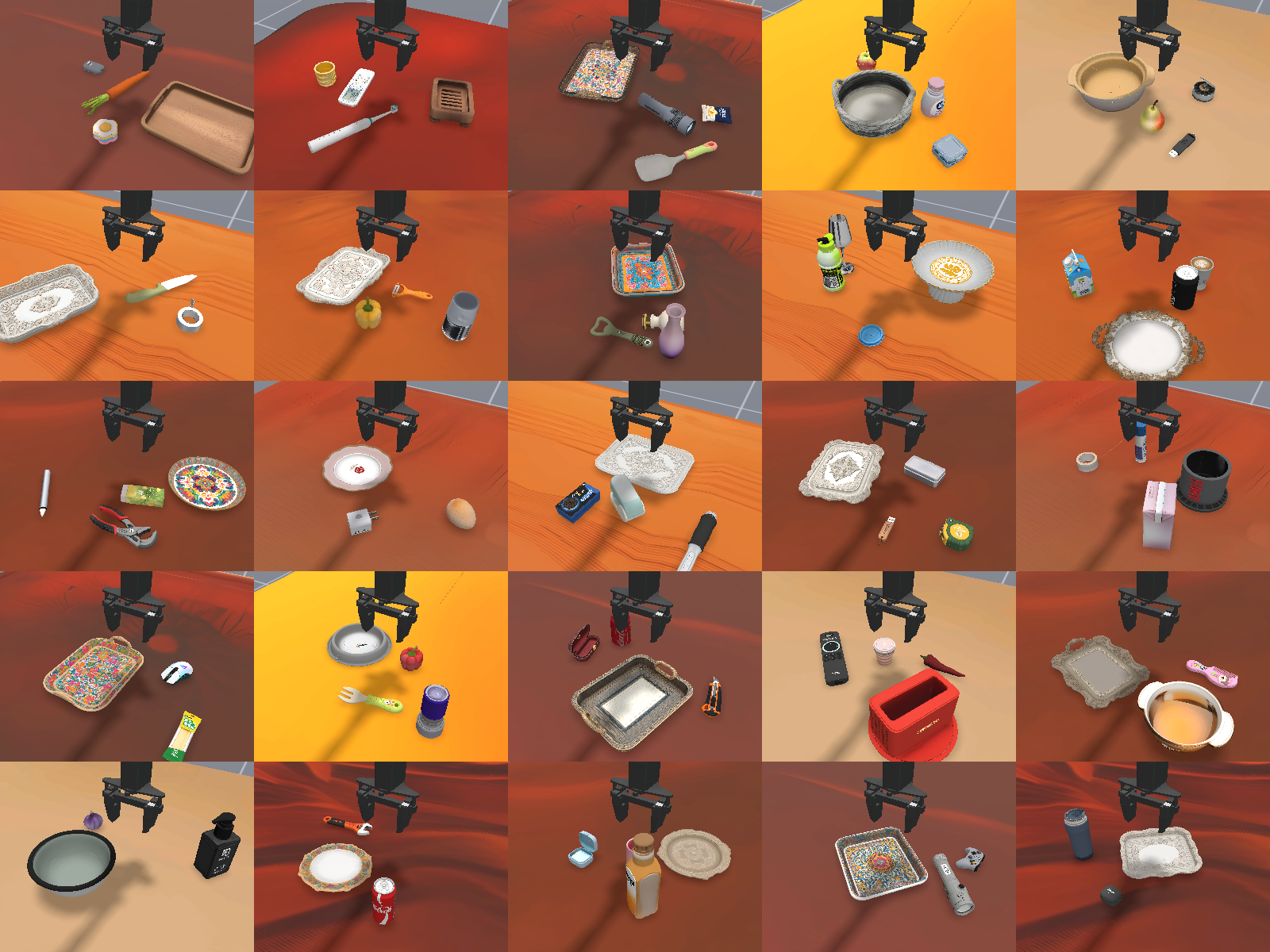}
\caption{
Hunyuan3D-2.1 scenes shown with domain randomization disabled. Evaluation was performed with domain randomization enabled.
}
\end{figure}

\begin{table}[H]
\renewcommand{\arraystretch}{1.1}
\fontsize{6.5pt}{7.3pt}\selectfont
\caption{Hunyuan3D-2.1 generated scenes 0--24.}
\centering
\begin{tabular}{l l l | l l l}
\toprule
Scene & Language Command & Distractors
& Scene & Language Command & Distractors \\
\midrule

$0$  & carrot $\rightarrow$ tray
& sticky notes, earbud case
& $13$ & green measuring tape $\rightarrow$ tray
& USB drive, staples box \\

$1$  & toothbrush $\rightarrow$ soap dish
& mouthwash cup, AC remote
& $14$ & glue stick $\rightarrow$ round container
& juice, Scotch tape \\

$2$  & spatula $\rightarrow$ tray
& flashlight, potato chips
& $15$ & white mouse $\rightarrow$ tray
& cheese sticks \\

$3$  & apple $\rightarrow$ basket
& pill organizer, chewing gum
& $16$ & green fork $\rightarrow$ plate
& chili, stamp \\

$4$  & pear $\rightarrow$ bowl
& staples box, USB drive
& $17$ & glasses case $\rightarrow$ tray
& Coke can, cutter knife \\

$5$  & knife $\rightarrow$ tray
& Scotch tape
& $18$ & black remote $\rightarrow$ red holder
& capsicum, spice jar \\

$6$  & orange peeler $\rightarrow$ tray
& stamp, chili
& $19$ & pink comb $\rightarrow$ tray
& bowl \\

$7$  & bottle opener $\rightarrow$ tray
& diffuser bottle, vase
& $20$ & purple garlic $\rightarrow$ bowl
& hand sanitizer \\

$8$  & yellow bottle $\rightarrow$ tray
& wine glass, coaster
& $21$ & red can $\rightarrow$ tray
& wrench \\

$9$  & milk carton $\rightarrow$ tray
& Coke can, coffee cup
& $22$ & orange bottle $\rightarrow$ tray
& pill organizer, pen holder \\

$10$ & green eraser $\rightarrow$ tray
& presenter clicker, pliers
& $23$ & flashlight $\rightarrow$ tray
& gamepad \\

$11$ & brown egg $\rightarrow$ plate
& USB charger
& $24$ & earbud case $\rightarrow$ tray
& thermos \\

$12$ & white stapler $\rightarrow$ tray
& biscuit box, wrench
& & & \\

\bottomrule
\end{tabular}
\label{tab:third_party_generated_tasks_and_scenes}
\end{table}

\newpage

\section{Sim-to-real Scenes}
\label{sec:sim_to_real_scenes}

\begin{figure}[h]
\centering
\includegraphics[width=0.9\columnwidth]{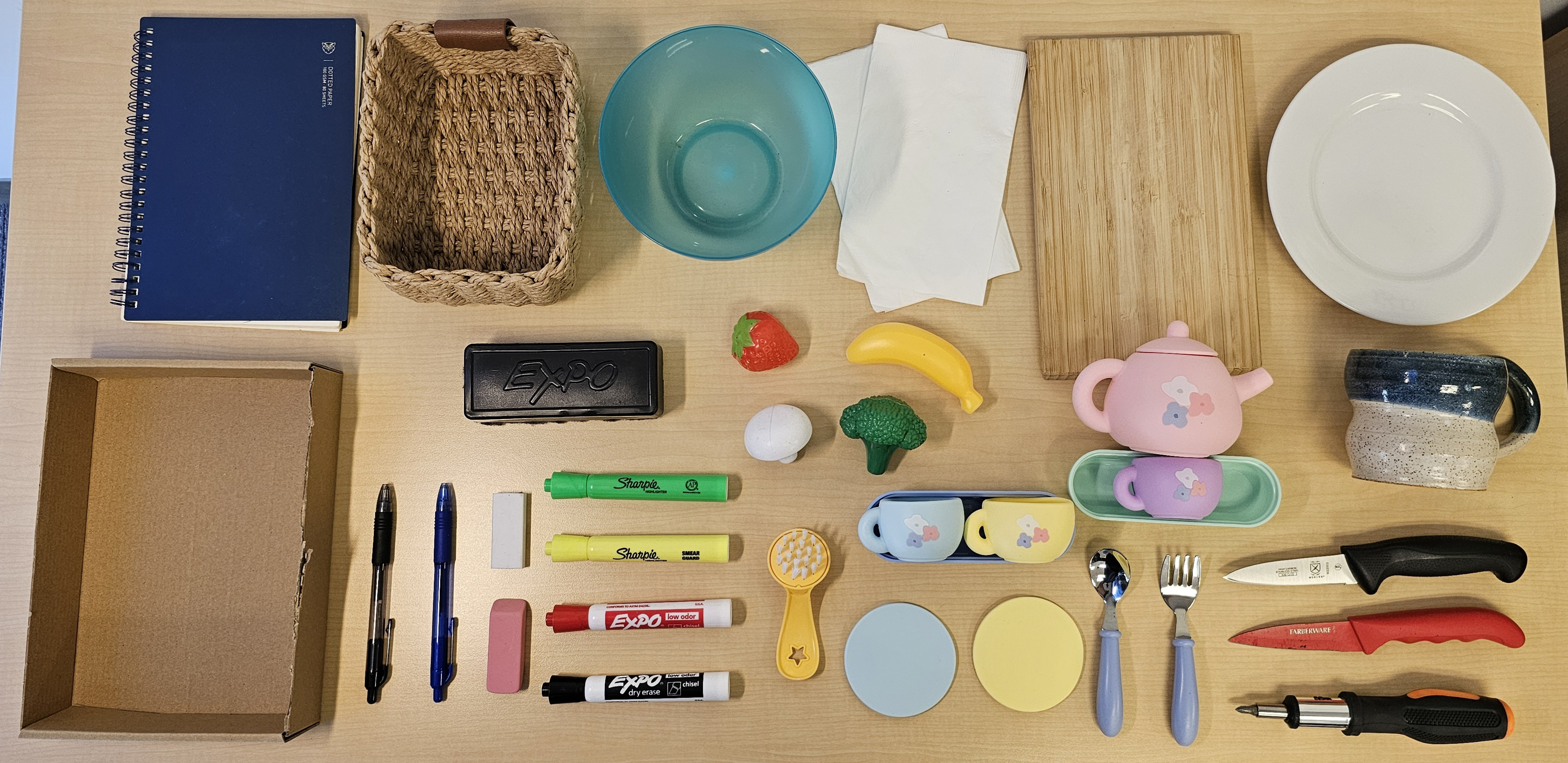}
\caption{
All objects used in the sim-to-real experiments.
}
\end{figure}

\begin{figure}[H]
\centering
\includegraphics[width=0.65\columnwidth]{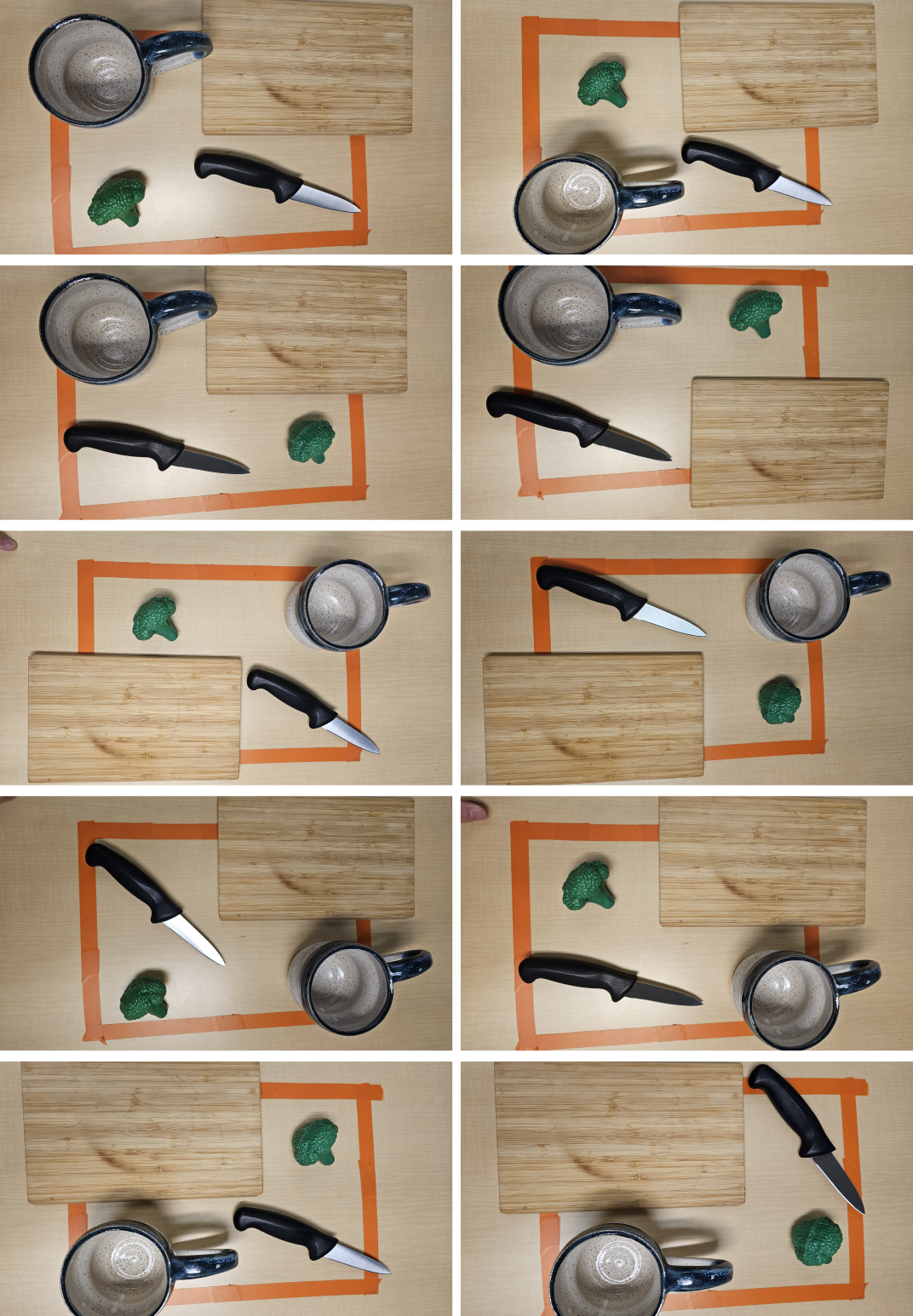}
\caption{
Example of 10 trial object randomization (scene 1 in Table~\ref{tab:sim_to_real}).
}
\end{figure}

\newpage

\section{Real-world Rollout Examples}
\label{sec:real_world_rollouts}

\begin{figure}[H]
\centering
\includegraphics[width=\columnwidth]{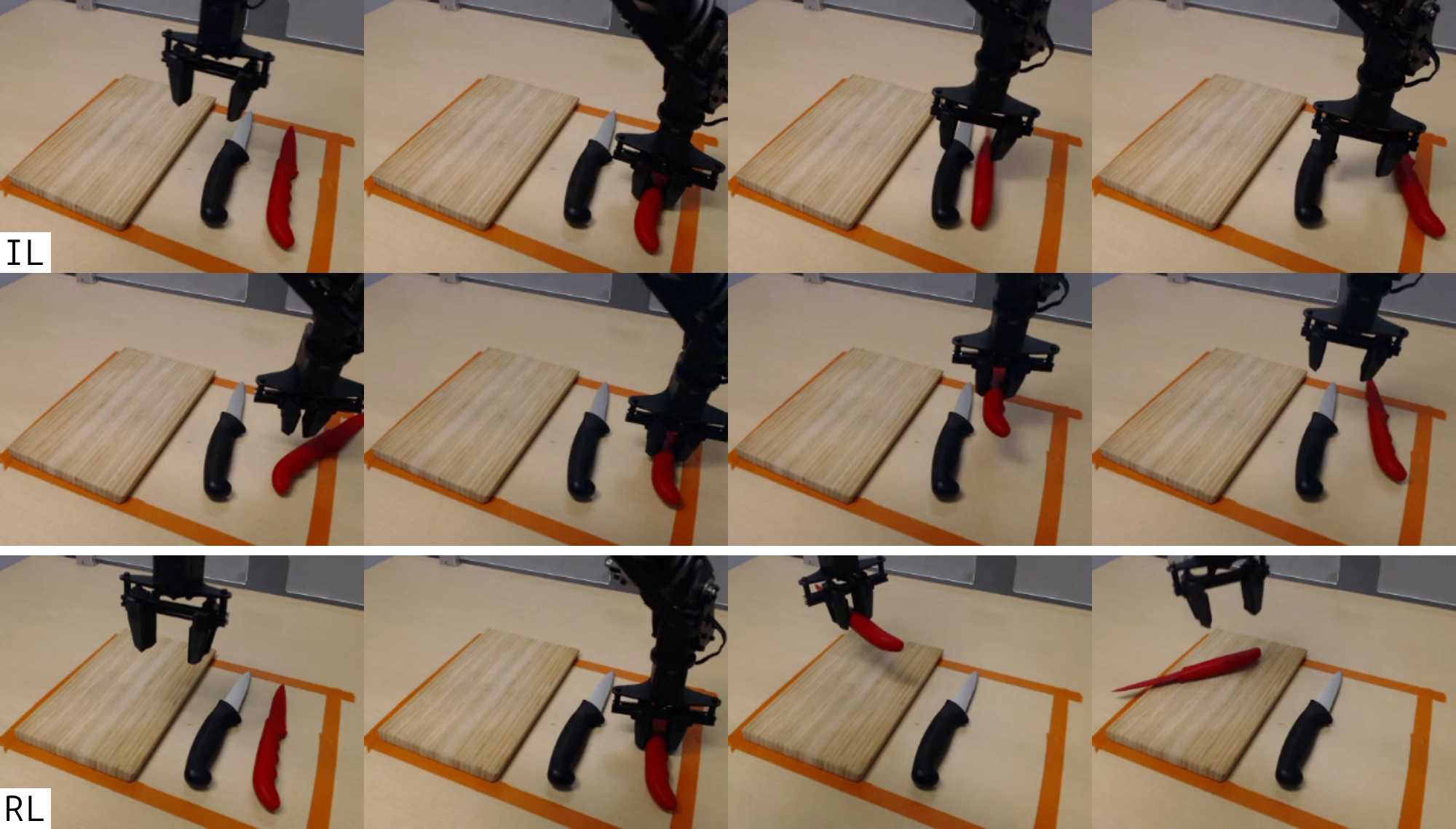}
\caption{
\textbf{IL: semantic failure. RL: success.}
A trial example of the red knife $\rightarrow$ cutting board task (scene~7 from Table~\ref{tab:sim_to_real}).
The imitation learning (IL) policy first grasps the red knife but immediately drops it after entering an out-of-distribution state.
It then grasps and lifts the knife once more before dropping it again and eventually timing out.
In comparison, the reinforcement learning (RL) policy grasps the red knife and deliberately places it on the cutting board.
}
\label{fig:red_knife_frames}
\end{figure}

\begin{figure}[H]
\centering
\includegraphics[width=\columnwidth]{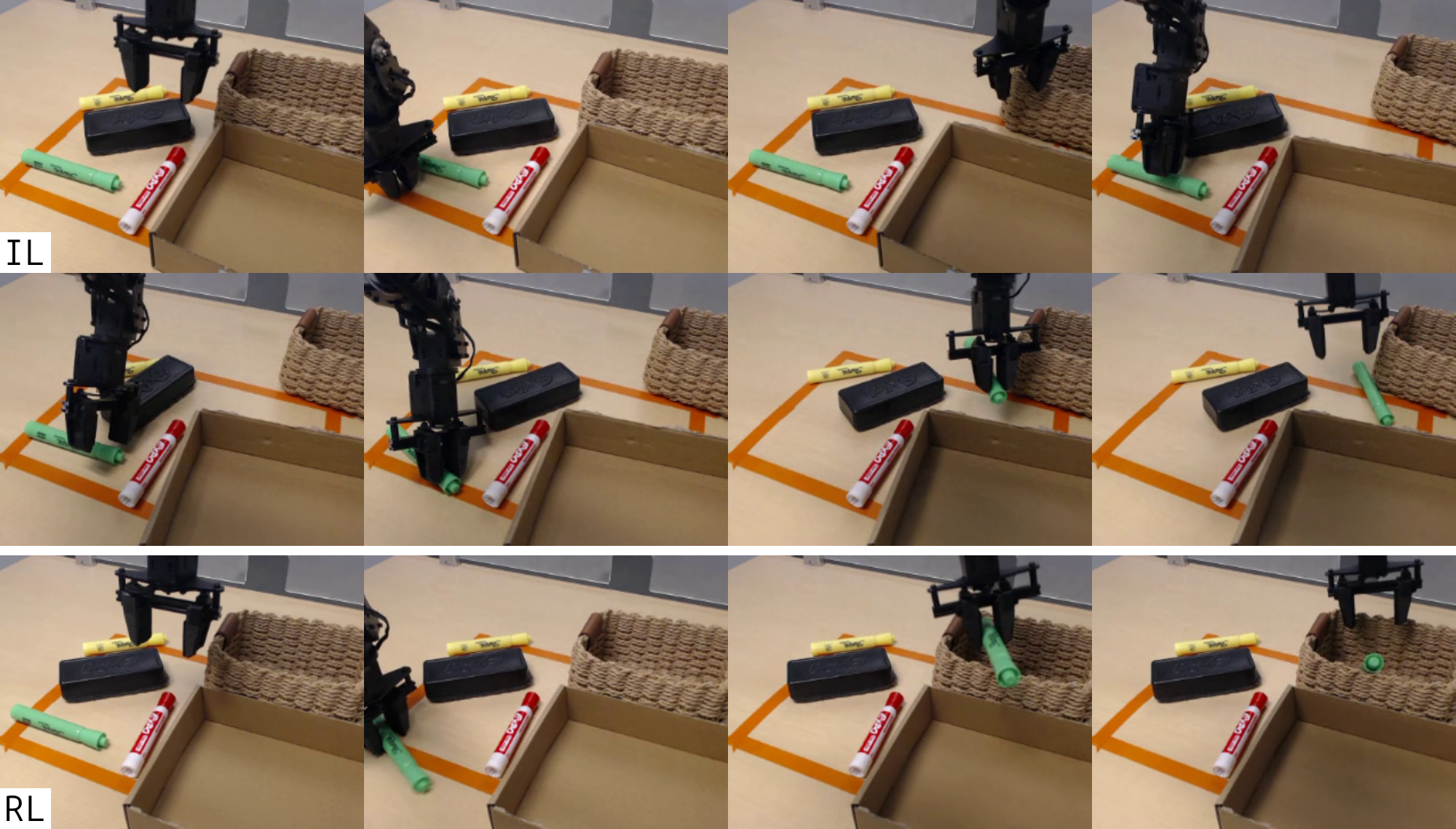}
\caption{
\textbf{IL: dynamics failure. RL: success.}
A trial example of the green marker $\rightarrow$ basket task (scene~9 from Table~\ref{tab:sim_to_real}).
The IL policy initially misses the grasp on the green marker and then pushes the basket away.
It makes two regrasp attempts before finally lifting the marker, but places it next to the basket instead and eventually times out.
In comparison, the RL policy correctly places the green marker inside the basket.
}
\label{fig:green_marker_frames}
\end{figure}

\newpage

\begin{figure}[H]
\centering
\includegraphics[width=\columnwidth]{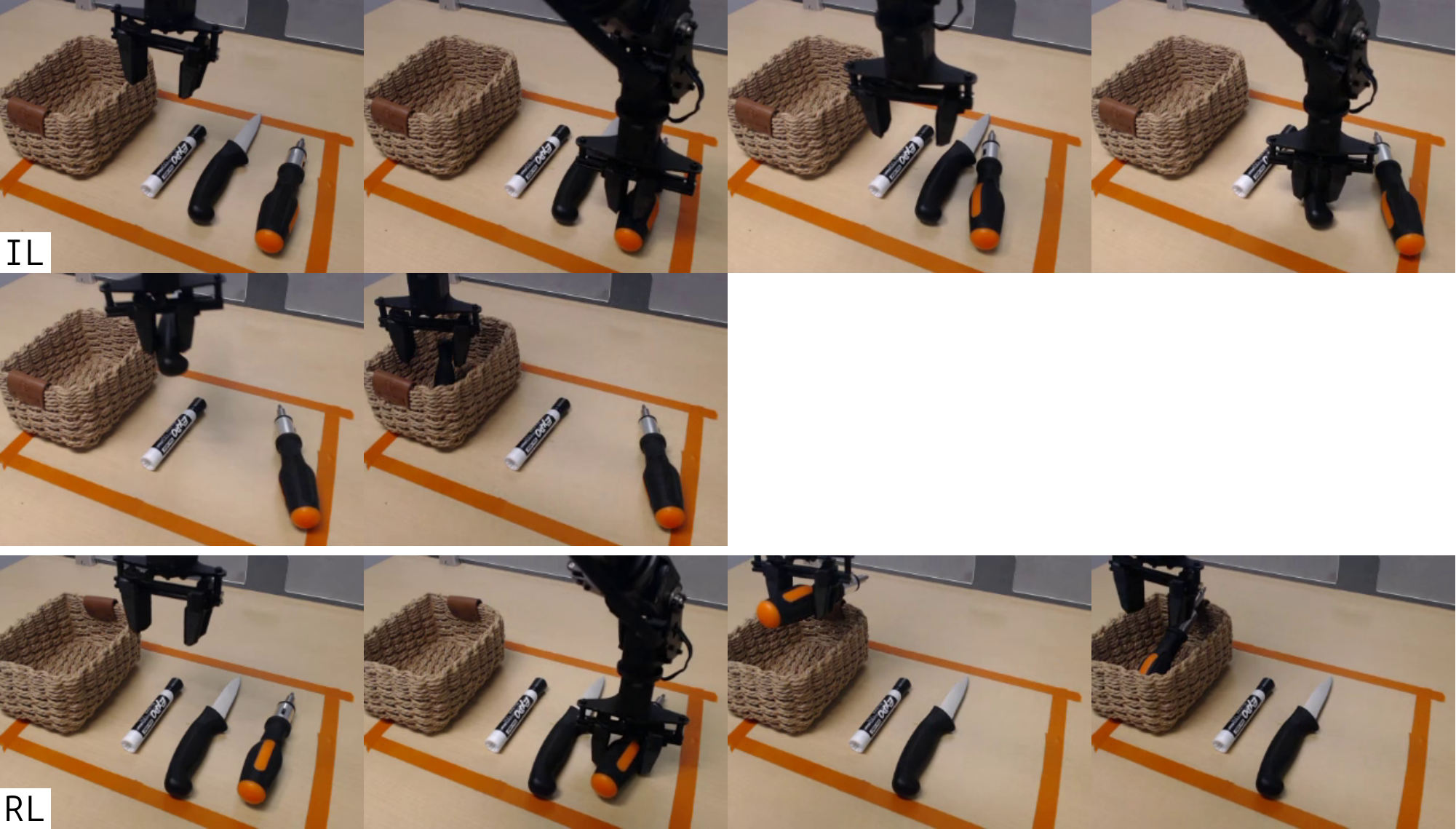}
\caption{
\textbf{IL: dynamics and semantic failure. RL: success.}
A trial example of the screwdriver $\rightarrow$ basket task (scene~10 from Table~\ref{tab:sim_to_real}).
The IL policy attempts to grasp the screwdriver but misses.
It then becomes confused and grasps the knife instead, placing it in the basket and failing due to incorrect task execution.
In comparison, the RL policy correctly places the screwdriver in the basket.
}
\label{fig:screwdriver_frames}
\end{figure}

\begin{figure}[H]
\centering
\includegraphics[width=\columnwidth]{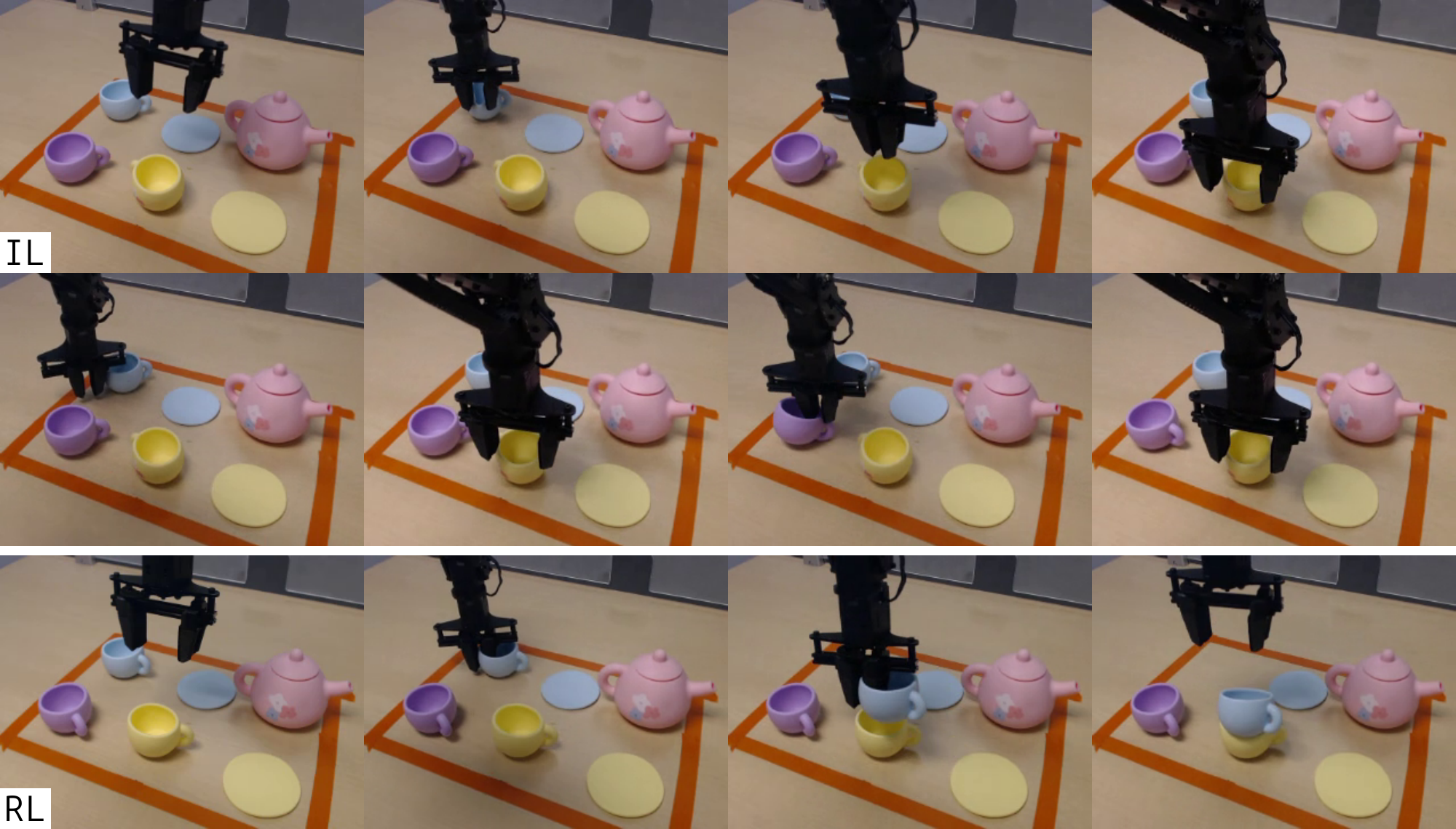}
\caption{
\textbf{IL: dynamics and semantic failure. RL: success.}
A trial example of the blue teacup $\rightarrow$ yellow teacup stacking task (scene~11 from Table~\ref{tab:sim_to_real}).
The IL policy first misses the grasp on the blue teacup and then moves toward the yellow teacup twice.
It subsequently becomes confused and attempts to grasp the purple teacup instead, eventually timing out.
In comparison, the RL policy quickly grasps the correct teacup and stacks it onto the yellow one.
}
\label{fig:teacup_stack_frames}
\end{figure}

\newpage

\begin{figure}[H]
\centering
\includegraphics[width=0.9\columnwidth]{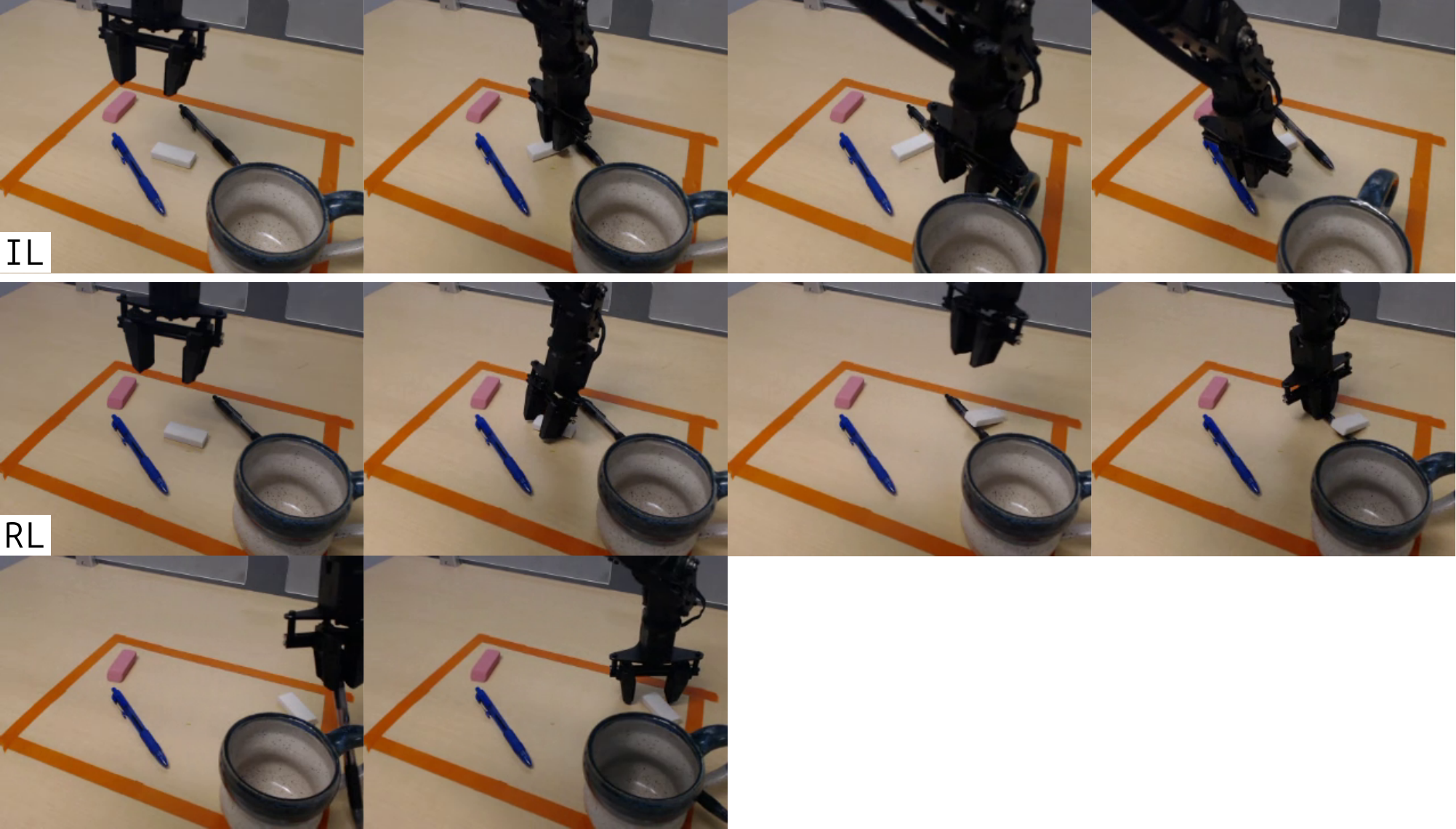}
\caption{
\textbf{IL: dynamics failure. RL: dynamics failure.}
A trial example of the white eraser $\rightarrow$ mug task (scene~6 from Table~\ref{tab:sim_to_real}).
The IL policy misses the grasp on the white eraser several times before ultimately hovering, leading to a timeout.
In contrast, the RL policy successfully grasps and lifts the white eraser, but it slips from the grasp and falls onto the black pen, confusing the policy.
The policy then grasps and clears the black pen before returning to the white eraser, but still times out.
}
\label{fig:white_eraser_frames}
\end{figure}

\begin{figure}[H]
\centering
\includegraphics[width=0.9\columnwidth]{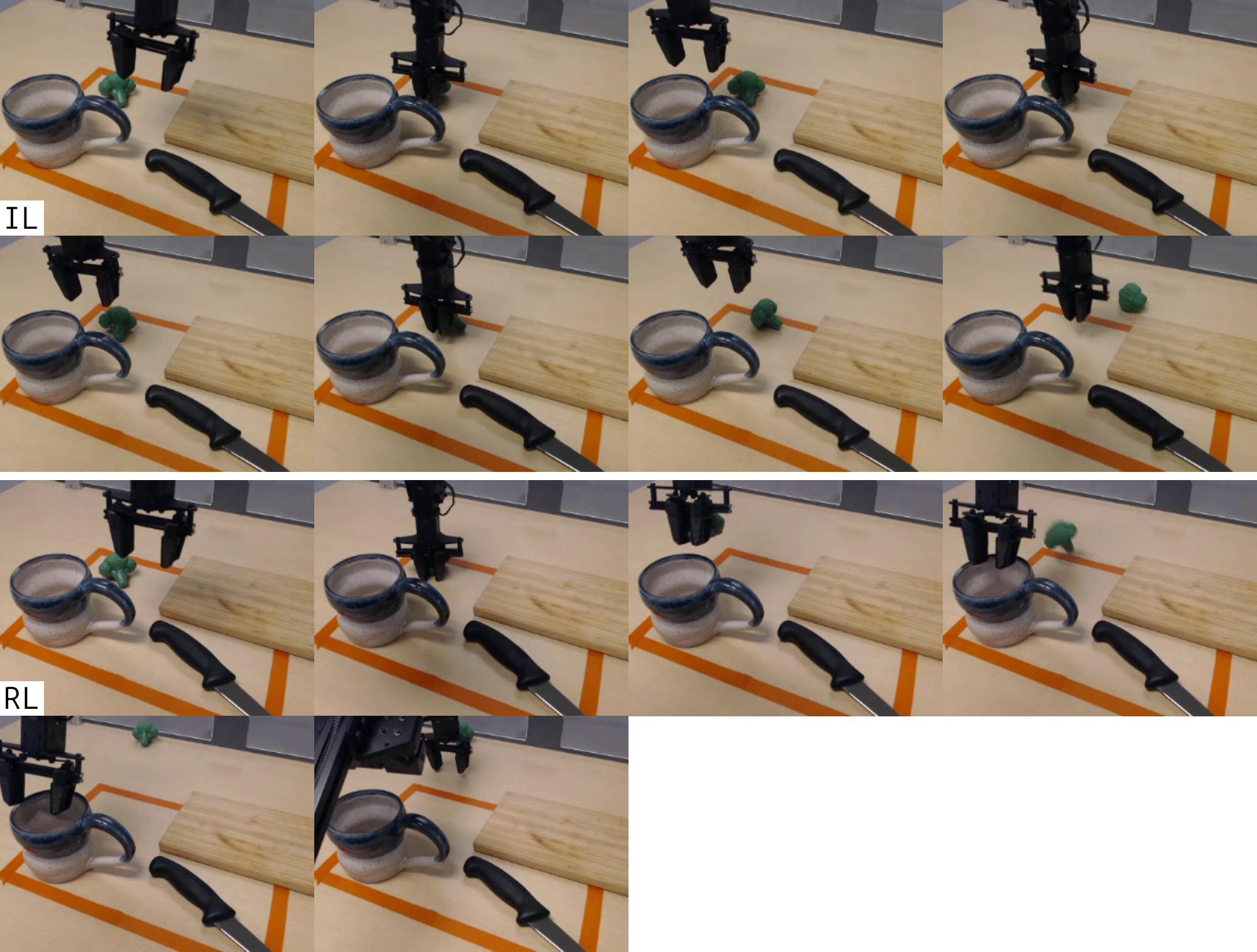}
\caption{
\textbf{IL: dynamics failure. RL: dynamics failure.}
A trial example of the broccoli $\rightarrow$ mug task (scene~1 from Table~\ref{tab:sim_to_real}).
The IL policy repeatedly misses the grasp on the broccoli and eventually times out.
In contrast, the RL policy quickly grasps the broccoli and attempts to place it into the mug, but it hits the edge and rolls out of reach of the manipulator.
A significant portion of failures was caused by objects moving out of reach, preventing regrasp attempts.
}
\label{fig:broccoli_frames}
\end{figure}

\end{document}